\documentclass[11pt]{article}

\usepackage[final]{acl}

\usepackage{times}
\usepackage{latexsym}
\usepackage{booktabs}
\usepackage{tabularx}
\usepackage{amsmath}
\usepackage{enumitem}
\usepackage{multirow}
\usepackage[table]{xcolor}
\usepackage[T1]{fontenc}

\usepackage[utf8]{inputenc}

\usepackage{microtype}

\usepackage{inconsolata}

\usepackage{graphicx}
\usepackage{subcaption}

\title{Probing Negation Understanding in Vision-Language Models With Object Attention}

\author{
 \textbf{Beth Pearson\textsuperscript{1}},
 \textbf{Michael Wray\textsuperscript{1}},
 \textbf{Martha Lewis\textsuperscript{2}}
\\
 \textsuperscript{1}University of Bristol,
 \textsuperscript{2}ILLC, University of Amsterdam
\\
 \small{
\texttt{\{beth.pearson, michael.wray\}@bristol.ac.uk, m.a.f.lewis@uva.nl}
}
}

\begin{document}
\maketitle
\begin{abstract}
Negation is a fundamental aspect of language understanding and it is important vision-language models (VLMs) are able to process it effectively. 
We investigate how VLMs process negation by analysing attention patterns of different models. We draw comparisons between human gaze and attention in models when processing affirmative vs negated statements. We compare object level model attention to human eye-fixations in both binary and multary settings, evaluating with both synthetic and realistic datasets. We find models generally do not follow human-like attention patterns when resolving negation, suggesting they rely on mechanisms outside of attention. While layerwise analysis shows that PaliGemma2-3B shifts object attention to the alternative under negation, our headwise ablations suggest this is a byproduct rather than a negation processing mechanism. Additionally, we find that LLaVA-1.5-7B displays a spatial bias towards left-side objects. Our evaluation of models on downstream tasks shows that all models have a discrepancy between affirmative and negated questions highlighting the need to better understand the underlying processing of negation to close this performance gap. Code available at \href{https://github.com/otmive/negation_attention}{github.com/otmive/negation\_attention}
\end{abstract}

\section{Introduction}
Despite their strong performance across various multi-modal tasks, vision-language models (VLMs) still struggle with negation. Investigating the ability of models to process negation is an emerging field with benchmarks to test negation accuracy being proposed (e.g. \citet{bui_nein_2025, zhang2025negvqa}) and explorations into how models such as CLIP handle negated captions~\cite{alhamoud2025vision, quantmeyer2024and}. Comparison with human performance on these benchmarks indicates that model performance is generally lower \cite{zhang2025negvqa}, in some cases very substantially \cite{kamath_scale_2026}.

At the same time, negation has been argued to be more difficult for humans to process \cite{wason_response_1961, clark_process_1972}, and more recently,~\citet{greco2020modulating}, with negated phrases taking a longer time to process than their affirmative counterparts. The cause of these longer reaction times is still under debate~\cite{nieuwland2008truth, kaup2006processing}. Given a negated sentence $\neg S$, e.g. \textit{the door is not open}, the delay in processing such a sentence could be accounted for by needing to apply a symbolic negation operator to $S$, or by needing to simulate the alternative affirmative equivalent to  $\neg S$, i.e. \textit{the door is closed}.

One approach to understanding how humans process negation is via the \textit{visual world paradigm}, which presents verbal and visual inputs simultaneously. \citet{orenes_how_2014} use this paradigm to investigate whether affirmative alternatives are simulated in both binary and multary contexts, and find that while humans do indeed simulate the affirmative in binary contexts, but not in multary contexts.

Since many of the errors that VLMs make in processing negation is to overpredict the affirmative $S$ \cite{alhamoud2025vision,zhang2025negvqa}, it is important to understand any parallels between VLM and human negation processing.

We examine possible parallels between attention to image tokens in VLMs and gaze fixation in humans. Using visual stimuli similar to those in \citet{orenes_how_2014}, we analyse the attention values of generated output back to the input image for three VLMs, each in two different sizes: LLaVA-1.5-7B/13B, InternVL3.5-4B/8B, and PaliGemma2-3B/10B. We create a synthetic dataset for evaluating models in Binary and Multary settings, similar to the human experiment setup in ~\cite{orenes_how_2014}. We also compare with attention values for realistic images more similar to the models' pre-training data by using images from the What'sUp~\cite{kamath_whats_2023} dataset. Our main contributions are threefold. First, we create two datasets for evaluating VLM negation processing in Binary (two object) and Multary (four object) settings, and generate bounding box annotations for the What'sUp dataset to facilitate object-specific attention analysis. Second, we evaluate object-level attention for 6 VLMs and assess the extent to which attention patterns in VLMs parallel human gaze fixations for affirmative and negated prompts. We find that in general, attention patterns in tested VLMs do not parallel human gaze fixations, with the exception of PaliGemma2-3B. Third, we demonstrate that attention to relevant objects in a scene does not necessarily correlate with strong downstream performance. As such, we contribute to the literature on the understanding of VLM negation processing.

\section{Background}
\paragraph{Negation in Vision-Language Models}
Recent studies show that VLMs struggle with handling negation~\cite{zhang2025negvqa, alhamoud2025vision, vatsa2025no, park2025know, singh2025learning}. Many VLMs suffer from affirmation bias because they are pre-trained on large datasets of images labelled by humans who tend to focus on positive concepts in the image e.g. (``a photo of a dog'') so models learn to associate the presence of words with the presence of objects~\cite{ettinger2020bert}. Negated phrases tend to appear in similar contexts to their affirmative counterparts, meaning that learned vectors of affirmative (`a dog') and negated (`not a dog') phrases can be extremely similar.
Overall, negation processing in VLMs is weak and further understanding is important for the development of more reliable models. 

\paragraph{Negation Processing in Humans}
Exactly how humans process negation is an open question, but it is agreed that negation is also harder for humans to process than affirmative statements~\cite{greco2020syntax, orenes2021looking}, evidenced by the fact that it takes longer for humans to process~\cite{kaup2006processing, clark_process_1972, greco2020modulating, wason_response_1961}. One theory of negation processing is that negation is handled using a two-step process: first the affirmative version of the statement is simulated and then negated in the second step~\cite{kaup2006processing}. Other theories argue that negation is processed incrementally, more similar to how affirmative sentences are processed~\cite{anderson2010temporal, tian2010we}.

The focus of our paper is on data collected by \citet{orenes_how_2014}. In that paper, an eye-tracking experiment is performed to investigate how negation is processed in humans. They use 2D images of coloured shapes with 4 shapes in an image. Each participant is presented with one of these images and 2 spoken sentences. In the binary experiment, the sentence mentions two possible colours eg. ``The figure could be red or blue''. The second sentence is then either affirmative: ``The figure was blue'' or negated: ``The figure was not blue''. The multary experiment considers four colours: ``The figure could be red, green, blue or yellow''. In the binary case, eye-gaze is initially on the mentioned colour and shifts after a short time delay to the alternative colour as the negation is processed. In the multary case, however, the gaze stays focused on the mentioned colour. These results are argued to provide evidence that in humans, in the binary case, negation is processed by simulating the affirmative alternative (i.e., ``the figure was red''). This is inferred from the focus on the non-blue object. On the other hand, in the multary case, it is argued that negation may be represented symbolically. This is inferred from the focus on the blue object, with the assumption that some kind of mental negation operator is applied to a mental representation of the blue object. Since VLMs suffer from an affirmation bias, it is important to understand what kinds of negation processing may be occurring.

\paragraph{Attention Visualisation}
Attention visualisation has been used in many works as an interpretability tool for VLMs to see which part of an image models are ``looking'' at when processing a specific word or query~\cite{ben2024lvlm, aflalo2022vl, yu2024understanding, bi2025unveiling}. Investigations into how CLIP processes negation find that there is a lack of separation between affirmative and negated captions~\cite{alhamoud2025vision} and that CLIP processes negation using specific heads at early and late layers~\cite{quantmeyer2024and}. 
However, there is less work looking at models not based on CLIP, and in how attention differs across different negation setups. 
In this paper, we extend investigations to a more diverse range of models and to how attention to mentioned and alternative objects is affected by different image set-ups. 
While the exact role of attention in text and image processing is still open to debate \cite{xing2025large}, evidence suggests that the information for an object is localised to the token positions corresponding to its original location in the image. Attention from these tokens in late to middle layers carries important object information to the final token of an input, and successful spatial reasoning correlates with the alignment of attention to tokens corresponding to object locations in images \cite{neo_towards_2025, chen_why_2025}. 
This suggests model attention patterns may be used to infer that the model is processing information from that specific region of the stimulus, similar to how human eye-gaze patterns may be used to identify the area the stimulus being processed~\cite{orenes_how_2014}. Additionally, integrating human gaze data to VLM input during training has been shown to boost performance~\cite{yan2024voila}. These points suggest that while model attention cannot serve as a direct proxy for human eye fixations, comparing model attention with human gaze is an interesting avenue for research.

\section{Datasets}

While the original experiment tracked the temporal movements of human eye-gaze, our work focuses on attention in VLMs processing negation, which is static. ~\citet{orenes_how_2014} use a two-sentence set-up where participants hear a phrase such as ``The figure could be red or blue'' (binary) or ``The figure could be red, blue, green, or yellow'' (multary) concurrently with the image, followed by a target sentence. LLaVA and PaliGemma models accept only inputs of the form (image, text), i.e. we cannot provide a stimulus of the form (text, image, text). We therefore replicate the original priming effect in a static domain using images containing either two shapes or four shapes, mirroring the number of alternatives introduced in the original experiment. Despite the difference in experimental setup, the processing of one alternative vs multiple (three) alternative objects remains the same. 

\paragraph{Binary}
We generate images containing a plain white background and two 2D objects. We consider the colours blue, red, green and yellow and the shapes circle, square, triangle and diamond.
Evaluating the models with these constraints ensures that this information will be in-domain for the vision-language models and aligns our experiments with the human eye-tracking experiments in \citet{orenes_how_2014}.
Objects within an image are of different colours but always the same shape and size. Between images, objects may be one of three sizes: small, medium or large. We choose values 0.12, 0.18, and 0.25 as the size values which act as shape diameter multipliers. For instance, a large circle will have a diameter of exactly 25\% of the image width. Positions of the objects, i.e. the coordinates of the centres of the shapes are fixed. Object sizes and colours are randomly assigned and we generate 1000 images in total. This ensures there are at least 100 images containing each possible colour-shape combination allowing sufficient evaluation of the models in all cases. 

\begin{figure}[htbp]
     \centering
     \begin{subfigure}[b]{0.45\columnwidth}
         \centering
         \includegraphics[width=\textwidth]{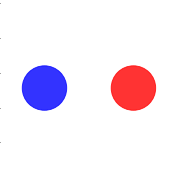}
         \caption{Binary dataset example}
         \label{fig:2_shape_ex}
     \end{subfigure}
     \hfill
     \begin{subfigure}[b]{0.45\columnwidth}
         \centering
         \includegraphics[width=\textwidth]{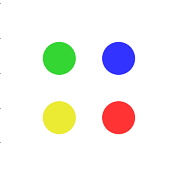}
         \caption{Multary dataset example}
         \label{fig:4_shape_ex}
     \end{subfigure}
     
     \vspace{1em} 
     
     \begin{subfigure}[b]{0.45\columnwidth}
         \centering
         \includegraphics[width=\textwidth]{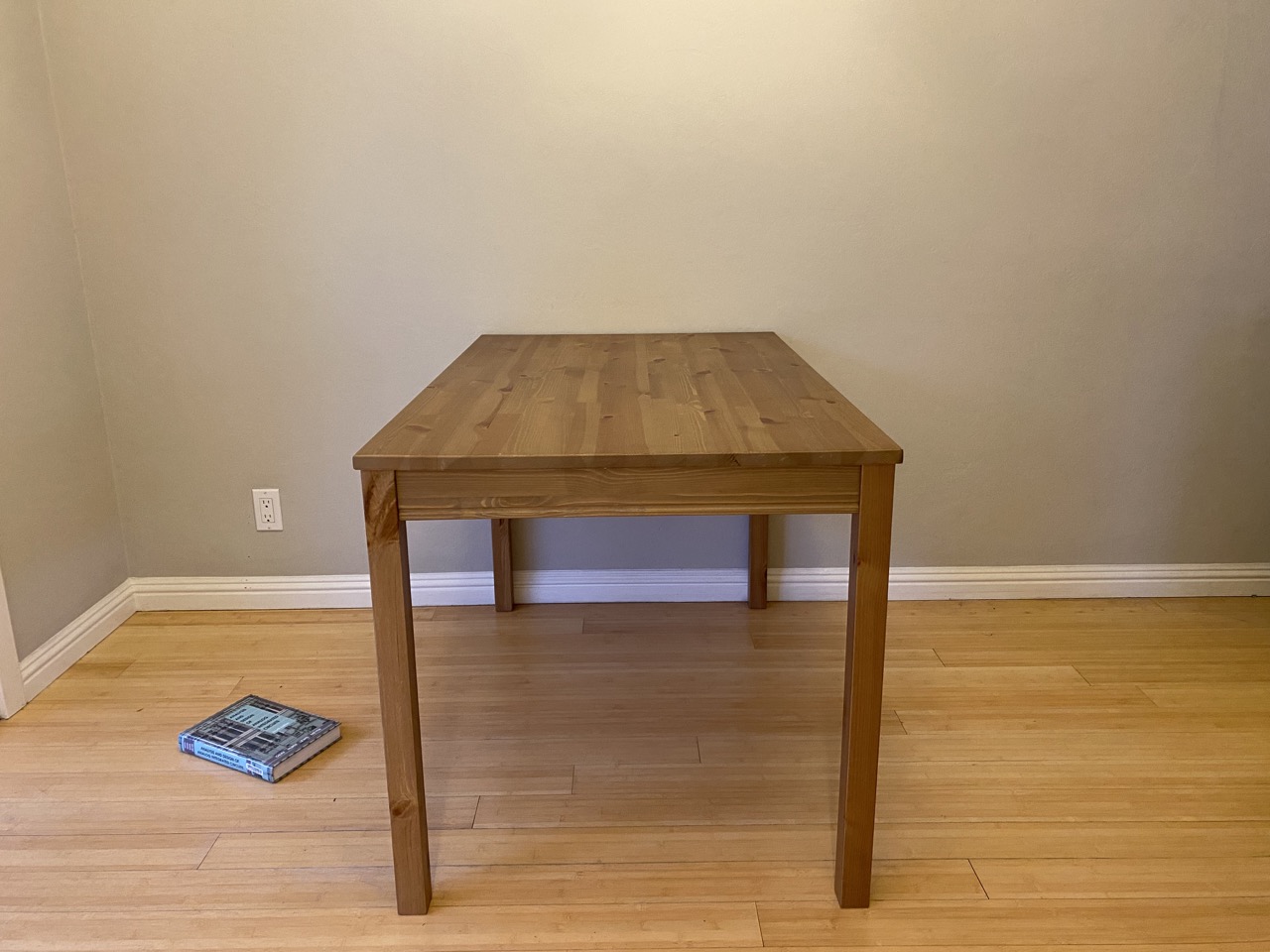}
         \caption{What'sUp: book left of table}
         \label{fig:whatsup_book_table}
     \end{subfigure}
     \hfill
     \begin{subfigure}[b]{0.45\columnwidth}
         \centering
         \includegraphics[width=\textwidth]{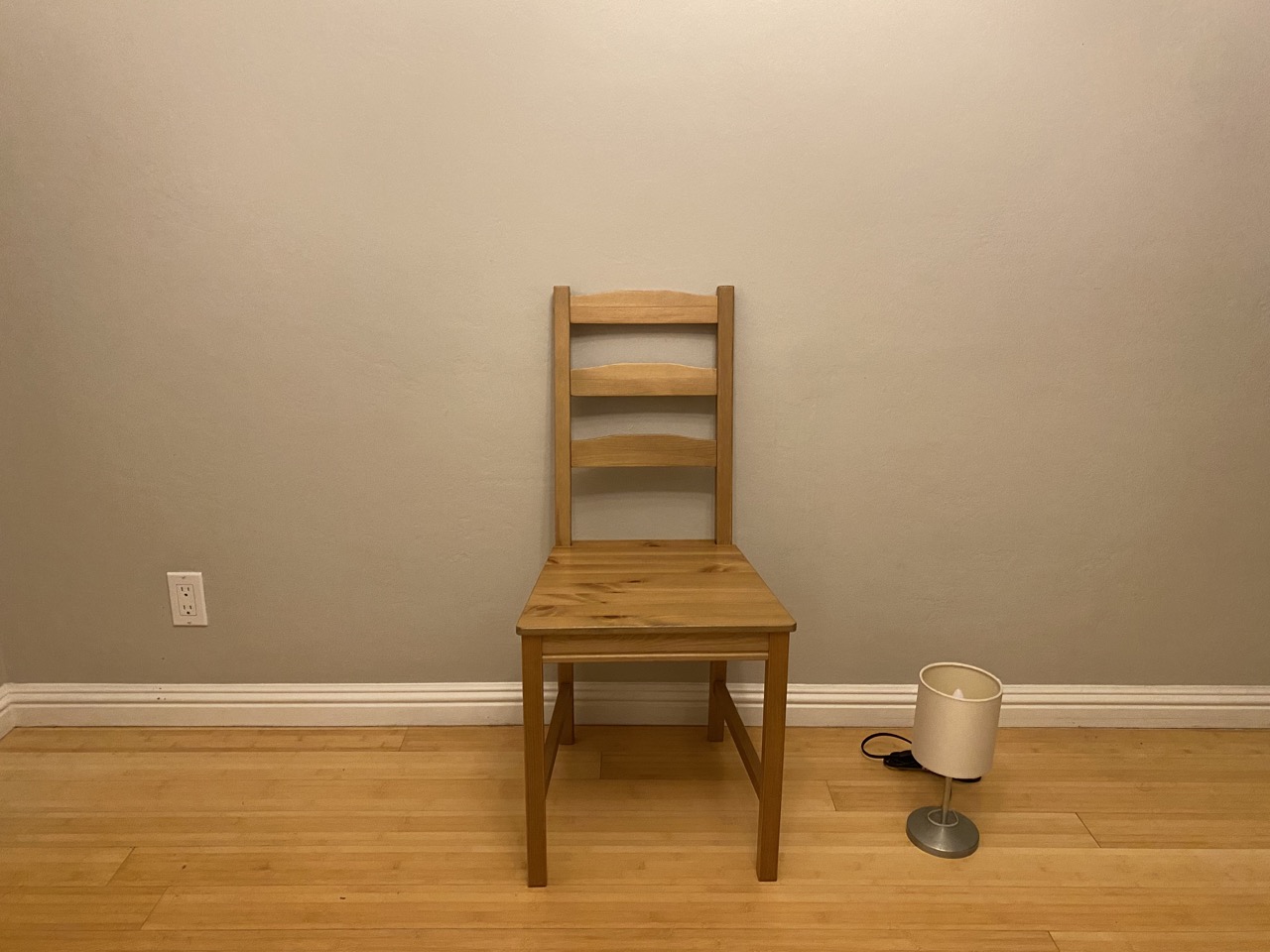}
         \caption{What'sUp: lamp right of chair}
         \label{fig:whatsup_lamp_chair}
     \end{subfigure}
     
     \caption{Dataset example images: (a) and (b) show images from the Binary and Multary datasets, while (c) and (d) show examples from the What'sUp dataset.}
     \label{fig:dataset_images}
\end{figure}

\paragraph{Multary}
Additionally, we generate images with four 2D objects on a plain white background. We consider the same colours and shapes as in the Binary dataset however we choose different size parameters 0.10, 0.14 and 0.18 for small, medium and large respectively. This maintains a consistent proportion of white background in the images between the Binary and Multary datasets. Again we generate 1000 images with fixed shape positions. Each image contains one object of each colour. Figure \ref{fig:dataset_images} shows an example from the Binary and Multary datasets.
\paragraph{What'sUp~\cite{kamath_whats_2023}}
We choose the What'sUp dataset as a realistic set of images to test whether the attention patterns for synthetic environments generalise to more complex real-world images. Given VLMs are primarily pre-trained on real-world datasets, attention mechanisms may be more optimised for realistic images. What'sUp contains images of two objects in four positions: `left', `right'. `on' and `under'. We exclude the relations `on' and `under' to avoid instances where objects occlude each other. Furthermore, this makes our attention analysis directly comparable with the horizontal layout of the Binary dataset. After we have applied the above filtering, our dataset uses 206 What'sUp images (out of 408), examples of which are shown in Figure~\ref{fig:dataset_images}.

To analyse attention to objects within an image, we use bounding boxes to map image tokens to corresponding objects. The Binary and Multary datasets contain bounding box annotations from our dataset generation. However, since the What'sUp dataset does not include bounding boxes, we set up a pipeline using Grounding DINO~\cite{liu2023grounding}. Specifically, we extract the entities of two target objects from each image using the objects within the ground truth label. Grounding DINO returns a list of candidate detections from which we selected those with the highest confidence and performed a qualitative evaluation finding these bounding boxes consistently aligned with object locations within images. Therefore, we select the candidate with the highest confidence score for each label to serve as our final bounding box annotations. To validate the accuracy of our pipeline, we manually annotated a subset of 20 images and calculated the Intersection over Union (IoU) between the model prediction and our ground truth. For a total of 40 objects (2 per image) the pipeline achieved an IoU of 0.93, demonstrating that the labels are suitable for our experiments. We provide a detailed breakdown in Appendix \ref{sec:iou}.

\section{Method}
\label{sec:method}
We analyse attention to objects in images together with statements that are either affirmative (e.g. ``the shape is \textcolor{blue}{blue}'') or negated (e.g. ``the shape is \underline{\textit{not} \textcolor{blue}{blue}}''). We use the following  terminology:
\begin{itemize}
\item \textbf{\textcolor{blue}{Mentioned}}: shape was mentioned in an affirmative sentence, e.g. a \textcolor{blue}{blue} shape given the sentence ``The figure is \textcolor{blue}{blue}''. 
\item \textbf{\textcolor{red}{Alternative}}: shape was not mentioned in an affirmative sentence, e.g. a \textcolor{red}{red} shape given ``The figure is \textcolor{blue}{blue}''. 
\item  \textcolor{blue}{\underline{\textbf{Mentioned Negated}}}: shape was mentioned in a negated sentence, e.g. a \textcolor{blue}{blue} shape given ``The figure is \underline{\textit{not} \textcolor{blue}{blue}}''.
\item \textcolor{red}{\underline{\textbf{Alternative Negated}}}: shape was not mentioned in a negated sentence, e.g. a \textcolor{red}{red} shape given ``The figure is \underline{\textit{not} \textcolor{blue}{blue}}''. 
\end{itemize}
Note how we use \textcolor{blue}{blue} to indicate a \textit{mentioned} shape, \textcolor{red}{red} to indicate an \textit{alternative} shape, and \underline{underlining} to indicate that the sentence is negated. This formatting is for clarity of presentation and was not present in the original human experiments.

\paragraph{Binary}
Inspired by the binary experiment in~\citet{orenes_how_2014}, we compare attention patterns to the two objects in the image in affirmative and negated contexts. If following a similar pattern to humans, 
we expect the following pattern of attention:
\begin{align*}
\textbf{\textcolor{blue}{Mentioned}} &> \textbf{\textcolor{red}{Alternative}} \\
\textcolor{red}{\underline{\textbf{Alternative Negated}}} &> \textcolor{blue}{\underline{\textbf{Mentioned Negated}}}
\end{align*}
Note that we do \emph{not} wish to compare the dynamics of gaze observed in humans, but rather the steady-state gaze behaviour observed after a few hundred milliseconds.

We compare the attention patterns for 3 model families: LLaVA-1.5~\cite{liu2024improved}, InternVL-3.5~\cite{wang2025internvl3} and PaliGemma2~\cite{Steiner2024PaliGemma2A}. For each model family, we evaluate 2 model sizes: LLaVA-1.5-7B and LLaVA-1.5-13B, InternVL-3.5-4B and InternVL-3.5-8B, and PaliGemma2-3B and PaliGemma2-10B. We selected these models based on their differences in model architectures and range of vision encoders and LLM backbones: LLaVA-1.5 (CLIP-ViT/Vicuna via MLP), InternVL3.5 (InternViT/Qwen3), and PaliGemma2 (SigLIP/Gemma-2). LLaVA processes images into 576 vision tokens, InternVL dynamically varies the number of image tokens based on input image size however, the number of visual tokens for InternVL is always 256 for images in our experiments, matching the fixed number of tokens for PaliGemma2. Table \ref{tab:model_comparison} shows a comparison of layer and head configurations. 
\begin{table}[h]
\centering
\small
\setlength{\tabcolsep}{3pt}
\resizebox{\columnwidth}{!}{%
\begin{tabular}{lcccc}
\toprule
\textbf{Model} & \textbf{Layers} & \textbf{Attn. Heads} & \textbf{Visual Tokens} \\ \midrule
LLaVA-1.5-7B & 32 & 32 & 576 \\
LLaVA-1.5-13B & 40 & 40 & 576 \\
InternVL3.5-4B & 36 & 32 & 256 \\
InternVL3.5-8B & 32 & 32 & 256 \\
PaliGemma2-3B & 26 & 8 & 256 \\
PaliGemma2-10B & 42 & 16 & 256 \\ \bottomrule
\end{tabular}
}
\caption{Configurations of VLMs}
\label{tab:model_comparison}
\end{table}

We analyse the cross-modal attention by capturing the weights assigned to image tokens during the generation of the output, which we limit to a single-token response. Here we focus on this single-token setting. Findings for multiple-token outputs are similar and reported in  Appendix \ref{sec:multi_token}. 
We use prompts that specify the colour of one of the image objects in either an affirmative or negated format e.g. ``The figure is \{colour\}'' or ``The figure is not \{colour\}''.

To analyse how different models process affirmative and negated inputs, we calculate normalised attention scores for each object within an image. For layer-wise attention we consider attention averaged over all heads.

We identify the image tokens based on the relevant marker for each model (e.g. $<\!image\!>$ for LLaVA models), and re-construct the 1D image token sequence to their original 2D coordinates. This results in an $n \times n$ grid of image attention weights for each transformer layer. To determine which image tokens correspond to specific objects, we project the bounding box data onto this grid. To map a coordinate from the image space $(W_{img}, H_{img})$ to the heatmap space $(n \times n)$, we use a scaling factor such that for any point $(x, y)$ in the original image, the transformed coordinates $(x', y')$ in the heatmap are:
\begin{equation}
    x' = \lfloor x \cdot \frac{n}{W_{img}} \rfloor, \quad y' = \lfloor y \cdot \frac{n}{H_{img}} \rfloor
\end{equation}
We then calculate normalised attention by summing attention within the bounding box and dividing by the number of tokens within the bounding box:
\begin{equation}
    Score = \frac{\sum_{j=y'_1}^{y'_2} \sum_{i=x'_1}^{x'_2} \text{attn}(i, j)}{(x'_2 - x'_1 + 1) \times (y'_2 - y'_1 + 1)}
\end{equation}
We report both layer-wise attention scores averaged across heads and mean attention for all layers and heads. 
Given that objects of different sizes will have a different number of tokens, we carry out an analysis of how object size impacts attention patterns which is provided in Appendix \ref{sec:obj_size}. Overall, we find that our results remain consistent across images containing varying object sizes.

\paragraph{Multary}
We repeat the Binary experiment with images containing 4 objects and compare with the multary experiment from~\citet{orenes_how_2014}. For humans presented with a negated sentence such as ``The figure is not blue'', rather than fixate on the alternative three shapes, fixations remain on the shape with the mentioned colour despite the negation. We therefore expect to see a reversed attention pattern for negated statements. That is, we expect to see:
\begin{align*}
\textbf{\textcolor{blue}{Mentioned}} &> \textbf{\textcolor{red}{Alternative}} \\
\textcolor{blue}{\underline{\textbf{Mentioned Negated}}} &>\textcolor{red}{\underline{\textbf{Alternative Negated}}}
\end{align*}

We calculate attention to the four objects in the image using the same method as in the Binary experiment. Mentioned attention is attention to the colour stated in the prompt, alternative attention is the mean of the attention to the other three objects. 

\paragraph{What's Up}
Finally, we repeat the Binary experiment using more realistic images. Since the models are pre-trained mainly on real-world data, they should have more familiarity with the objects in these images compared with the synthetic 2D stimuli. We use the same calculations as in the Binary experiment except prompts now take the form ``The object is a \{noun\}'' where \{noun\} is replaced with one of the two objects in the image.  Due to structure, the behaviour we expect to see is that of the Binary dataset.

\subsection{Downstream Tasks}
To test the relationship between attention patterns and downstream performance we evaluate models on all three datasets. We ask the models a binary question where the answer is either `Yes' or `No'. For each image, we consider four question categories: affirmative-true (A.T.): an affirmative question with the correct answer Yes; affirmative-false (A.F.): an affirmative question with the correct answer No;
negated-true (N.T.): a negated question with the correct answer Yes; and
negated-false (N.F.): a negated question with the correct answer No.
Our initial prompt format P1 is: ``Is this statement correct? The object on the \{position\} is \{colour\}. Answer yes or no.'' where 'position' can be `left' or `right' for binary and `top left', `top right', `bottom left' or `bottom right' for multary and 'colour' is chosen to align with each image. We repeat each question 5 times with different prompt variations and report the average. We use Gemini 3 Flash~\cite{gemini3flash} to generate prompt variations based on prompt template 1 (P1). We use the following prompt: ``\{P1\} can you generate differently worded variations of this prompt template'' and choose four alternatives from the produced list. A full list of prompt templates used for the downstream tasks is provided in Appendix \ref{sec:prompts}.

\section{Results}
\begin{figure*}[tbp]
    \centering
    \includegraphics[width=.9\textwidth]{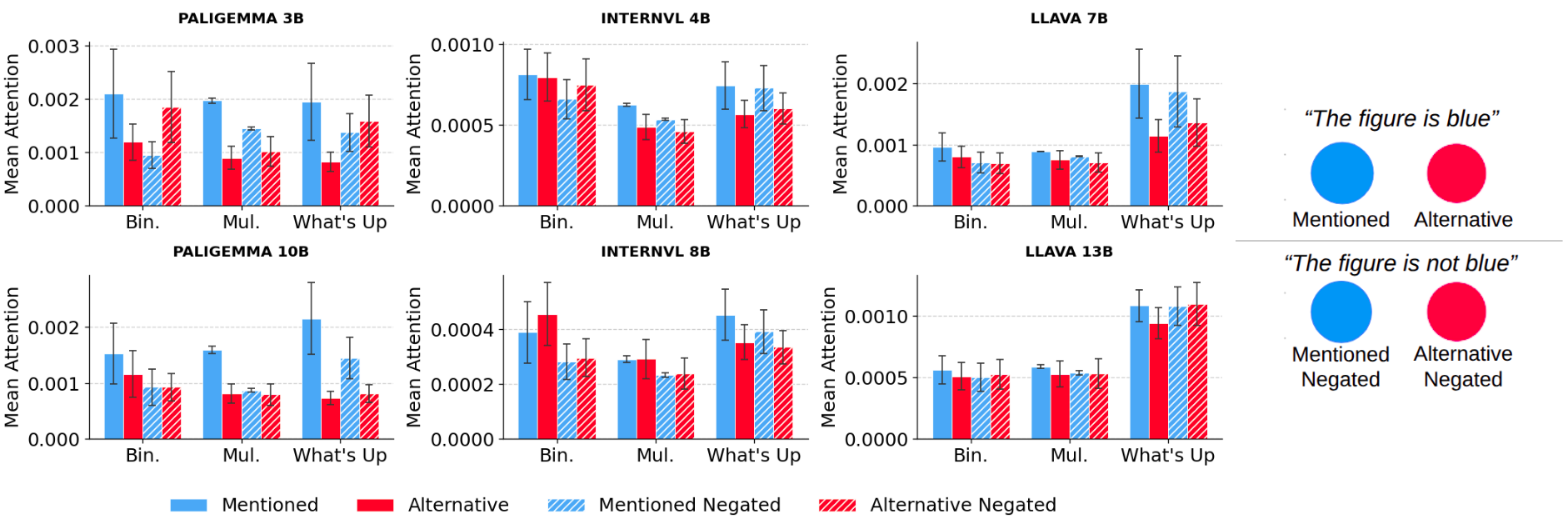}
    \caption{Mean attention to objects for models across datasets. For a human-like response we expect \textcolor{blue}{Mentioned} $>$ \textcolor{red}{Alternative} throughout, \textcolor{red}{\underline{Alternative Negated}} $>$ \textcolor{blue}{\underline{Mentioned Negated}} for the \textbf{Binary} and \textbf{What'sUp} datasets, and \textcolor{blue}{\underline{Mentioned Negated}} $>$ \textcolor{red}{\underline{Alternative Negated}} for the \textbf{Multary} dataset. Error bars represent the 95\% confidence interval.}
    \label{fig:bar_graph_results}
\end{figure*}

\subsection{High Level Results} Figure \ref{fig:bar_graph_results} shows the mean object attention for each dataset and model. Only PaliGemma2-3B shows the human-like pattern we outlined in section \ref{sec:method} across all datasets. InternVL-4B follows the expected pattern for Binary and Multary however on the What'sUp dataset more attention is always given to the mentioned object. InternVL-8B has more attention to the alternative shape in the affirmative Binary setting which is somewhat unexpected. Layerwise plots of attention to each object show that this is mainly due to a shift in attention occurring after layer 19 in which attention to the alternative object increases which occurs in for both affirmative and negated prompts. This phenomenon does not occur for the What'sUp dataset. LLaVA-7B and LLaVA-13B both have increased attention to the What'sUp dataset, suggesting that the more realistic objects in What'sUp, compared with those in the 2D datasets, influence where the models attend. Overall, we find that models generally do not exhibit attention patterns similar to those performed by humans in Binary and Multary negation contexts. This suggests models are not `simulating the alternative' but may be processing negation via other mechanisms. Additionally, there is large variation in attention behaviour between models indicating that attention is sensitive to model architectural configurations such as number of heads or layers. 

We calculated baseline attentions or the attention to visual tokens outside annotated objects for each model and dataset. Although attention sinks in irrelevant visual tokens can sometimes skew these values upward, we find that values are consistent across datasets (within $\pm 0.0001$). Mean baseline attention values for each model are: 0.0010 for PaliGemma-3B, 0.0009 for  PaliGemma-10B, 0.0010 for  InternVL-4B, 0.0011 for  InternVL-8B, 0.0006 for  LLaVA-7B, and 0.0007 for  LLaVA-13B.

To further evaluate our observations, we carried out Wilcoxon signed-rank tests to test the following hypotheses on the smaller model variants.
For the \textbf{Binary} and \textbf{What'sUp} datasets we hypothesise that \textcolor{blue}{M} $>$ \textcolor{red}{A} and \textcolor{red}{\underline{AN}} $>$ \textcolor{blue}{\underline{MN}}.
For the \textbf{Multary} dataset: \textcolor{blue}{M} $>$ \textcolor{red}{A} and \textcolor{blue}{\underline{MN}} $>$ \textcolor{red}{\underline{AN}}, where in each case the null hypothesis is that these inequalities do not hold.
We find that for PaliGemma2-3B, null hypotheses are rejected in all cases, and effect sizes are very large (Cohen's $d > 0.8$), with the exception of \textcolor{red}{\underline{AN}} $>$ \textcolor{blue}{\underline{MN}} on What'sUp with effect size Cohen’s $d$=0.17.
For InternVL-4B, null hypotheses are rejected for all except \textcolor{blue}{M} > \textcolor{red}{A} for binary and \textcolor{red}{\underline{AN}} $>$ \textcolor{blue}{\underline{MN}} for the What'sUp dataset, with generally small-medium effect sizes (Cohen’s d 0.2-0.45).
For LlaVa-7B, null hypotheses are not rejected for \textcolor{red}{\underline{AN}} $>$ \textcolor{blue}{\underline{MN}} on either the What'sUp or Binary datasets. Effect sizes are small-medium (Cohen’s d 0.25-0.6). 
Since the larger model variants generally deviate from the expected pattern, we do not carry out Wilcoxon signed-rank tests for these models.

\subsection{Layer-Wise Attention}
We now look in more detail at the attention to objects across different layers. Layer-wise attention to objects in the What'sUp dataset is shown in Figure \ref{fig:whatsup_results}. Layer-wise attention for the Binary and Multary datasets are provided in Appendices \ref{sec:binary_layer} and \ref{sec:multary_layer}. We do not draw parallels between layer-wise attention and temporal behaviour in humans, however. Layer-wise analyses are simply to shed light on VLM processes.

We see that in PaliGemma2-3B for negated prompts (second row, first column), there is a clear shift of attention from the mentioned object to the alternative at layer 13. LLaVA-1.5-10B has slightly increased attention to the alternative object between layers 16 and 25 but in layers 25+ attention remains similar to both objects. The other 4 models consistently allocate more attention to the mentioned shape than the alternative at all layers, even in binary negated contexts.

\begin{figure*}[bp]
    \centering
    \includegraphics[width=.9\textwidth]{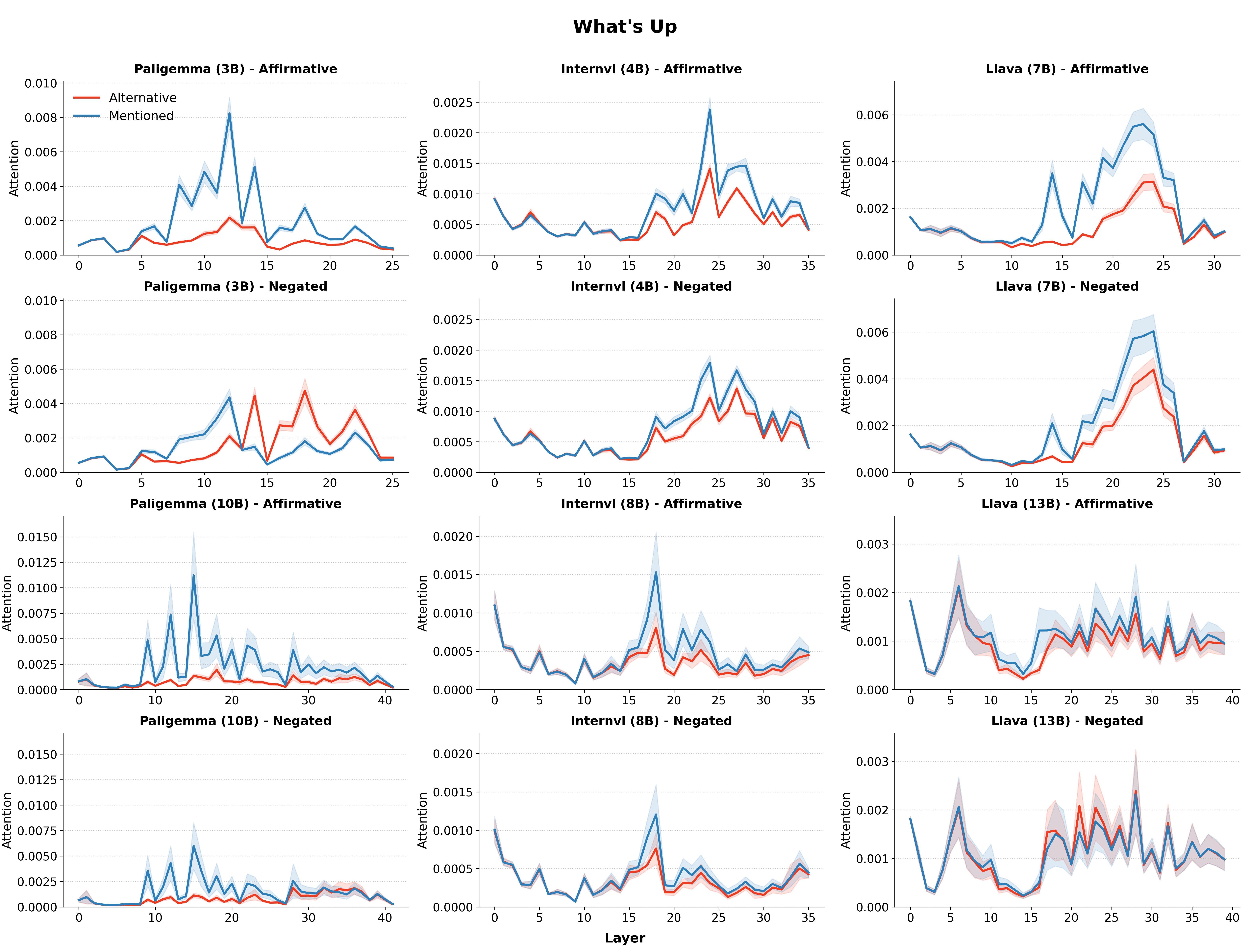}
    \caption{Layer-wise attention for What'sUp dataset}
    \label{fig:whatsup_results}
\end{figure*}

\subsection{Left-Right Bias}
\begin{figure*}[btp]
    \centering
    \begin{subfigure}[b]{0.9\textwidth}
        \centering
        \includegraphics[width=\textwidth]{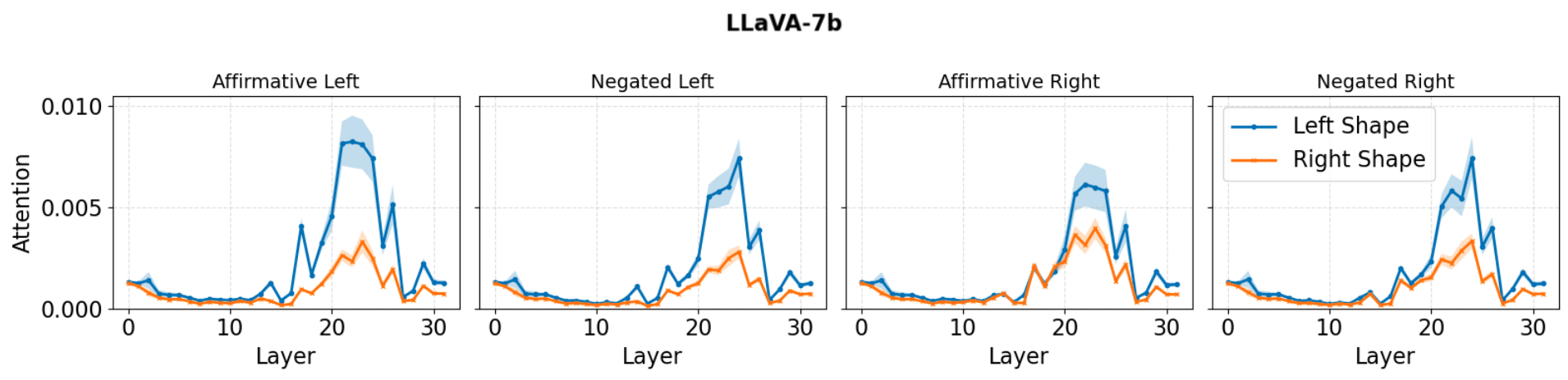}
    \end{subfigure}
    
    \vspace{-0.5em}

    \begin{subfigure}[b]{0.9\textwidth}
        \centering
        \includegraphics[width=\textwidth]{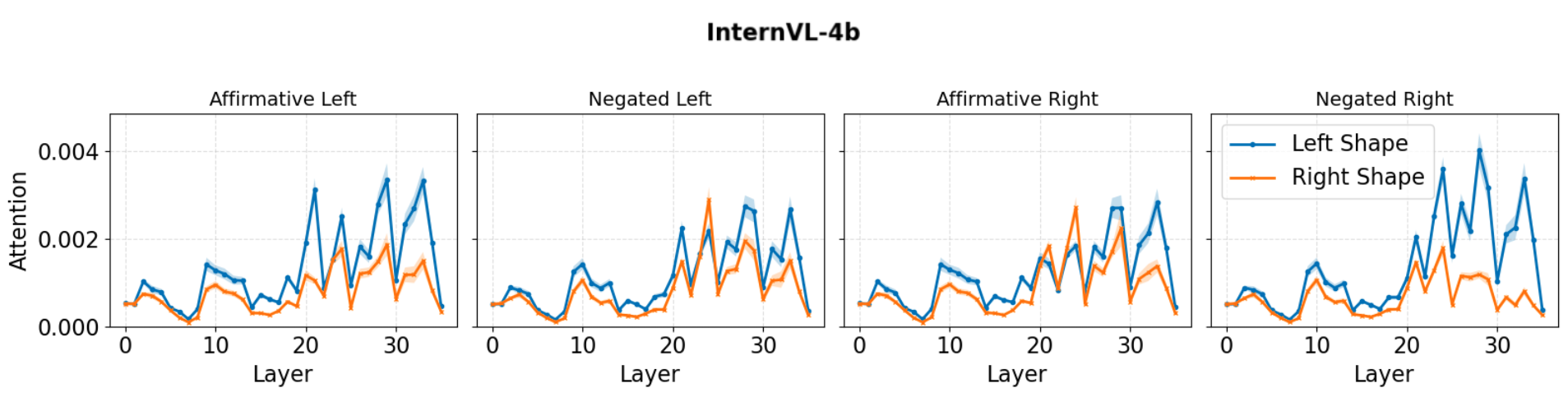}
    \end{subfigure}

    \vspace{-0.5em}

    \begin{subfigure}[b]{0.9\textwidth}
        \centering
        \includegraphics[width=\textwidth]{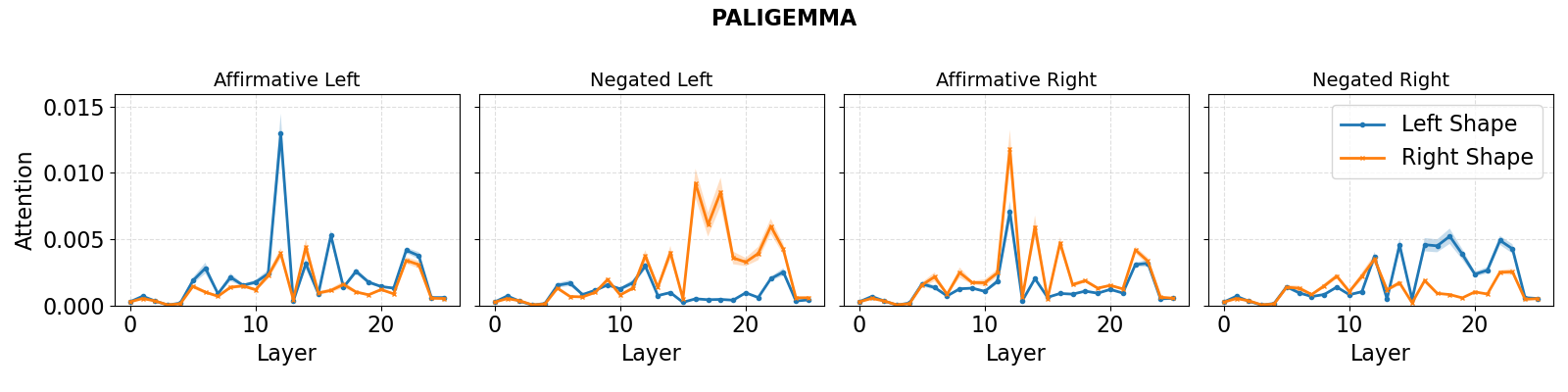}
    \end{subfigure}

    \caption{Layer-wise attention patterns for left-shape vs. right-shape colour mentions in both positive and negated contexts for smaller models.}
    \label{fig:left_vs_right_graphs}
\end{figure*}

To analyse the influence of spatial position, we plot layer-wise attention on the Binary dataset for smaller model sizes, categorising results by the position (left or right) of the mentioned object (Figure \ref{fig:left_vs_right_graphs}). Leftmost graphs correspond to the left object being mentioned, and rightmost graphs correspond to the right object being mentioned. We observe a bias in LLaVA-1.5-7B where the attention scores for the left object are consistently higher, especially at layers 20 to 25, regardless of which object is mentioned or whether the prompt is negated. There is slightly more attention to the right object in the affirmative right case than affirmative left, showing the prompt modulates LLaVA's focus to a certain extent. However, in the affirmative right case, attention to the unmentioned left object is still higher than the mentioned right object. While we see this effect slightly in InternVL-3.5-4B, there are still certain layers where the mentioned object has the most attention. For LLaVA-1.5-7B, the left shape consistently has the most attention suggesting LLaVA prioritises left-hand side visual tokens when generating its output. 

InternVL-3.5-4B shows a more even distribution of attention to the left and right objects. The attention patterns for affirmative left are most similar to the negated right and vice-versa. This suggests InternVL may be processing the negated prompts by converting them into their affirmative alternative, but that this is not happening in the attention weights. For example, given the Binary image in Figure \ref{fig:dataset_images}, ``The figure is \underline{not \textcolor{blue}{blue}}'' may be interpreted as ``The figure is \textcolor{red}{red}''. Comparing with the general attention pattern illustrated in Figure \ref{fig:bar_graph_results}, this suggests that the negation is not being actively computed in the attention weights, but may be calculated elsewhere in the model and causing attention to follow the pattern of the affirmative alternative.  In PaliGemma2-3B, we see that attention patterns for affirmative left are most similar to affirmative right and similarly for the negated left and right graphs, suggesting PaliGemma2-3B has the most symmetrical visual processing of the smaller models. 

\begin{table*}[tbp]
\centering
\small
\resizebox{\textwidth}{!}{%
\begin{tabular}{l cccc cccc cccc}
\toprule
 & \multicolumn{4}{c}{\textbf{Binary}} & \multicolumn{4}{c}{\textbf{Multary}} & \multicolumn{4}{c}{\textbf{What'sUp}} \\
\cmidrule(lr){2-5} \cmidrule(lr){6-9} \cmidrule(lr){10-13}
\textbf{Model} & \textbf{A.T} & \textbf{N.T} & \textbf{A.F.} & \textbf{N.F.} & \textbf{A.T} & \textbf{N.T} & \textbf{A.F.} & \textbf{N.F.} & \textbf{A.T} & \textbf{N.T.} & \textbf{A.F.} & \textbf{N.F.} \\ 
\midrule
LLaVA-7B & \textbf{100.0} & 22.7 & 53.6 & 99.3 & 93.5 & 45.5 & 31.7 & 73.9 & 88.7 & 3.5 &  29.0 & 95.7\\
LLaVA-13B & \textbf{100.0} & 39.5 & 79.1 & 88.5 & \textbf{100.0} & 66.5 & 24.6 & 51.7 & 85.1 & 22.7 &  40.7 & 86.3\\
 InternVL-4B & \textbf{100.0} & \textbf{100.0} & \textbf{100.0} & \textbf{99.8} & 92.7 & 77.1 & \textbf{99.9} & \textbf{97.7} & 83.3 & 69.9 & \textbf{75.8} & 78.5 \\
  InternVL-8B & \textbf{100.0} & 78.2 & 93.5 & 98.9 & 99.2 & \textbf{81.8} & 91.1 & 81.4 & 78.5 & 29.0 & 59.4 & 91.2 \\
 PGemma-3B & \textbf{100.0} & 76.2 & \textbf{100.0} & 96.6 &  \textbf{100.0} & 65.6 & 65.0 & 95.0 & \textbf{90.3} & 20.8 & 61.1 & \textbf{97.6} \\
  PGemma-10B & \textbf{100.0} & 54.4 & \textbf{100.0} & 45.7 &  \textbf{100.0} & 71.2 & 96.5 & 81.0 & 86.2 & \textbf{75.4} & 59.0 & 29.64 \\
\bottomrule
\end{tabular}%
}\caption{Downstream Accuracy For Models On All Datasets. We report mean accuracy for four question types: affirmative-true (A.T.), negated-true (N.T.), affirmative-false (A.F.) and negated-false (N.F.) }
\label{tab:downstream_results}
\end{table*}

\subsection{Downstream Tasks}

Table \ref{tab:downstream_results} shows the accuracy of models across four different categories of questions: affirmative-true (A.T.), negated-true (N.T.), affirmative-false (A.F.) and negated-false (N.F.). Each category type is run with 5 different seeds and across 5 different prompt templates. Full results available in Appendix \ref{sec:downstream_full}.

Downstream performance highlights a significant gap in performance between affirmative and negated prompts. Overall accuracies for all six models are lower for prompts containing negation than the affirmative prompts. There is however, large variation in accuracy between question categories, with models tending to get higher accuracies on A.T and N.F showing that question phrasing has a priming effect on models outputs, for example, the presence of the word `not' appears to bias questions towards `no'.

In humans, processing time is increased for negation however their overall error rate on simple attributes such as colour should be low. If VLMs had robust reasoning we would expect a high accuracy for both affirmative and negated questions. The drop in model performance on negated questions underscores the importance of understanding negation processing in VLMs and highlights the difficulty models have with handling negation.

PaliGemma2-3B has object-attention patterns most closely following the expected pattern from our headline results, however this doesn't translate into a stronger performance on downstream tasks. Intern-VL-3.5-4B achieves the highest overall performance despite showing similar layer-wise attention patterns for affirmative and negated prompts. In humans, processing negation may involve simulating the alternative (e.g. representing red for not blue). If a model were processing negation in a similar manner we would expect to see distinct differences in attention to objects when negation is introduced. Intern-VL-3.5-4B's strong results without such patterns suggest that human-like simulation is not required to successfully process negation. Instead, it points to the possibility that negation is processed more symbolically, separately to attention within the model's other components, allowing it to interpret negated prompts without visibly altering its attention. 

\section{Ablations and Headwise Analyses}
\begin{figure}[h]
    \centering
    \includegraphics[width=.45\textwidth]{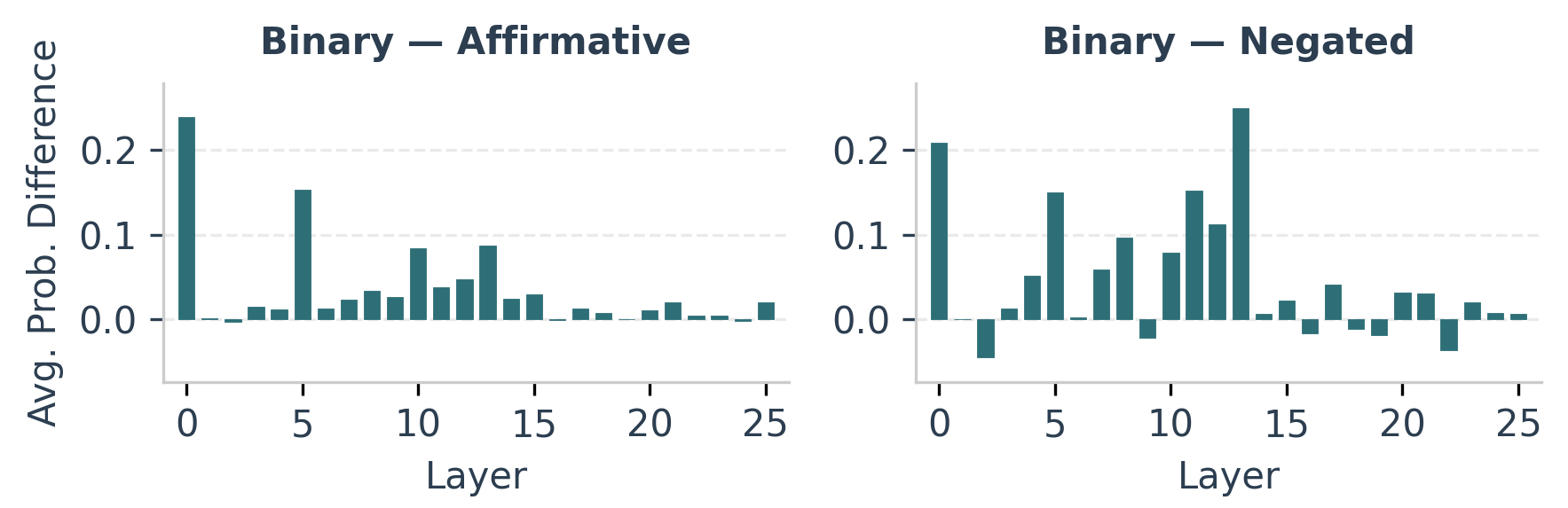}
    \caption{Effect layer ablations on binary dataset. A positive value means the targeted layer caused a decrease in correct prediction.}
    \label{fig:pali_ablation_binary}
\end{figure}
To further investigate the attention shift in PaliGemma2-3B, we perform a layer-wise ablation. Results are reported in Figure \ref{fig:pali_ablation_binary} and Appendix \ref{sec:ablation}. We find that ablating layers \textit{earlier} than layer 13 has the strongest causal effect on the output of the model. This provides evidence that in fact, information from the alternative is \emph{not} used in formulating the model's response. This is contrary to the conclusions from the gaze behaviour of humans. Instead, PaliGemma2-3B may process negation outside the attention mechanism in earlier layers, informing attention patterns for later layers. Additionally, comparing layer-wise attention in PaliGemma2-3B for correct and incorrect predictions shows that when making incorrect predictions, we don't see a shift in attention to the alternative (see Appendix \ref{sec:correct_incorrect}), suggesting flawed negation processing may be causing it to be unable to redirect its attention to the relevant visual features. As such, this forms an interesting avenue for further investigation.

We also report attention patterns for individual heads in Appendix \ref{sec:headwise}, Figure \ref{fig:total_heads}. We see that averaged across layers, the attention patterns of most heads are in line with the overall pattern. Exceptions are heads 0 and 7. Across heads and layers (Figure \ref{fig:layer_heads}), we see that individual heads generally show an overall attention shift to the alternative object in later layers. While we find no specialised `negation heads', different heads attend to the alternative object in different groups of layers -- some mid, and some later layers.

\section{Discussion and Conclusion}
We investigate negation understanding in VLMs by quantifying attention allocated to objects within images. We design experimental frameworks inspired by~\citet{orenes_how_2014} and compare our findings with human eye-gaze fixations patterns. In addition, we extend the What'sUp dataset \citep{kamath_whats_2023} using object segmentation to provide bounding boxes, facilitating object-level attention analysis on real-world imagery. We evaluate six models (two sizes across three model families: PaliGemma2, InternVL3.5 and LLaVA-1.5) finding that models don't generally exhibit attention patterns in-line with human eye-fixations, suggesting models do not process negation via `simulating the alternative' as humans are though to do. We theorise that models may be processing negation in a more symbolic manner, relying on components beyond attention. While PaliGemma-2-3B exhibits human-like attention shifts toward negated objects in later layers, we posit that this is not a simulation of the affirmative alternative as seen in human gaze. Instead, ablation studies show that early-layer attention has more causal effect on negation processing. We additionally find that LLaVA-1.5-7B has a bias to left-side shapes which may be contributing to it having the lowest accuracy on downstream tasks. All models show a drop in downstream accuracy for negated prompts, underscoring the fundamental gap in VLM negation processing. Understanding the underlying mechanisms of negation processing is crucial for developing more robust negation-aware architectures for future models. 

\section*{Limitations}
We acknowledge the following limitations in our work. Our analysis is limited to three models, while selected for their differences in model architectures, our experiments could be extended to a wider range of models in future work. Additionally, our experiments use simplistic images, even the more naturalistic What'sUp datasets contains only two objects under positional constraints. Exploring how models negation handling changes when faced with more complex images would be an interesting avenue for future research. 

There are slight differences in our experiment set-up compared with the human experiments from \citet{orenes_how_2014}. While the original human experiments use images containing four shapes with auditory cues to prime the participant for a binary or multary context, our experiments replace this temporal priming with images containing either two or four shapes. While there is a difference in priming method, the conditions are designed to replicate the binary and multary contexts for negation processing and we expect them to provide an analogous effect. 

We focus our analysis on attention to image tokens as a proxy for human eye-fixations. However, exploring other mechanisms for example the residual stream may reveal more insights into how the models process negation.
Finally, we look only at negation in simple settings and in English. Exploring other languages is an area for future work.

\section*{Ethical Considerations}
This work contains no direct ethical risks as we rely on open-source model evaluations and publicly available human data from ~\citet{orenes_how_2014}. However, we would like to acknowledge the wider societal impact that our finding of weak negation handling may have. The discrepancy in performance on negated prompts which we observe across all models highlights a fundamental limitation in negation understanding. This has serious implications, especially in high-stakes domains such as medical, safety and legal applications where exclusion criteria are critical~\cite{alhamoud2025vision}. Understanding how VLMs process negation as this is essential for safe and reliable deployment in these domains. 

\section*{Acknowledgments}
The authors wish to acknowledge and thank the financial support of the UK Research and Innovation (UKRI) [Grant ref EP/S022937/1] (BP) and the University of Bristol (BP, MW). ML acknowledges financial support from the Sectorplan science fund of the Dutch government. This work was carried out using the computational facilities of the Advanced Computing Research Centre, University of Bristol---http://www.bristol.ac.uk/acrc/.

\bibliography{references}

\appendix

\section{Appendix}
\label{sec:appendix}

\subsection{Bounding Box Validation}
\label{sec:iou}
To ensure the accuracy of the bounding boxes calculated by Grounding DINO~\cite{liu2023grounding} on the What'sUp~\cite{kamath_whats_2023} dataset, we calculate the Intersection over Union (IoU) with manually annotated bounding boxes on a subset of 20 images. We randomly sampled 20 images from our What'sUp dataset, resulting in 40 objects (2 per image) for evaluation. A human annotator manually drew bounding boxes around objects for each image, using the same text labels as provided to Grounding DINO. For example, `apple' and `armchair' for the image `apple left of armchair'. We evaluated spatial overlap using the Intersection over Union (IoU) metric, defined as:$$IoU = \frac{\text{Area of Overlap}}{\text{Area of Union}}$$
The pipeline achieved an IoU of 0.93 with a standard deviation of 0.08 for the 40 objects annotated. The minimum IoU observed across our evaluation set was 0.69 which we note was for the object `apple' which occupies a particularly small region of the image. 

\subsection{Multiple Token Outputs}
\label{sec:multi_token}
\begin{figure*}[t]
    \centering
    \begin{subfigure}[b]{0.9\textwidth}
        \centering
        \includegraphics[width=\textwidth]{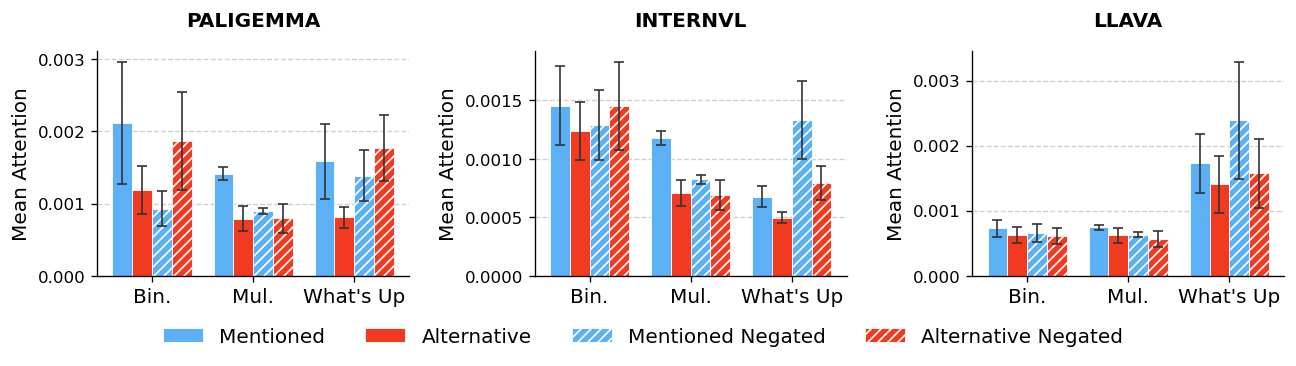}
        \caption{Two token output.}
        \label{subfig:same_size}
    \end{subfigure}
    
    \vspace{0.3cm}
    
    \begin{subfigure}[b]{0.9\textwidth}
        \centering
        \includegraphics[width=\textwidth]{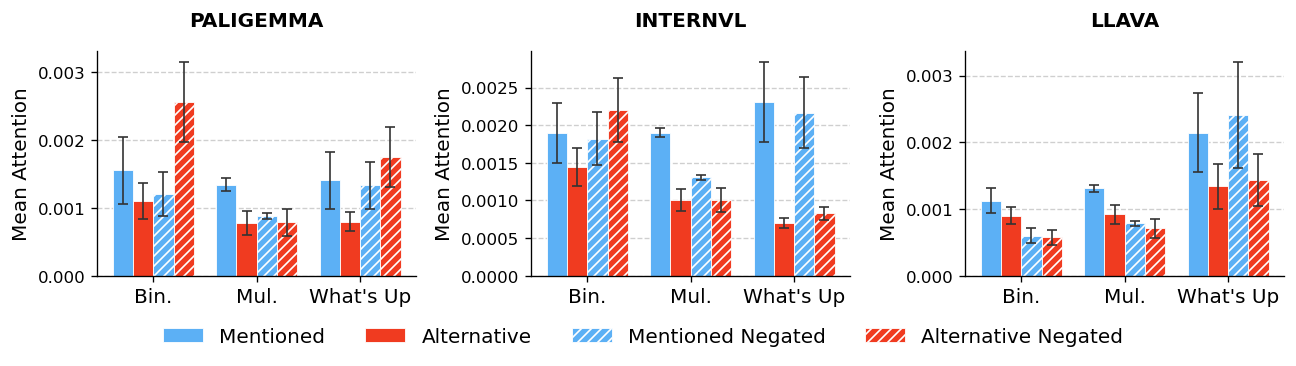}
        \caption{Five token output.}
        \label{subfig:size_left_large}
    \end{subfigure}
    
    \vspace{0.3cm}
    
    \begin{subfigure}[b]{0.9\textwidth}
        \centering
        \includegraphics[width=\textwidth]{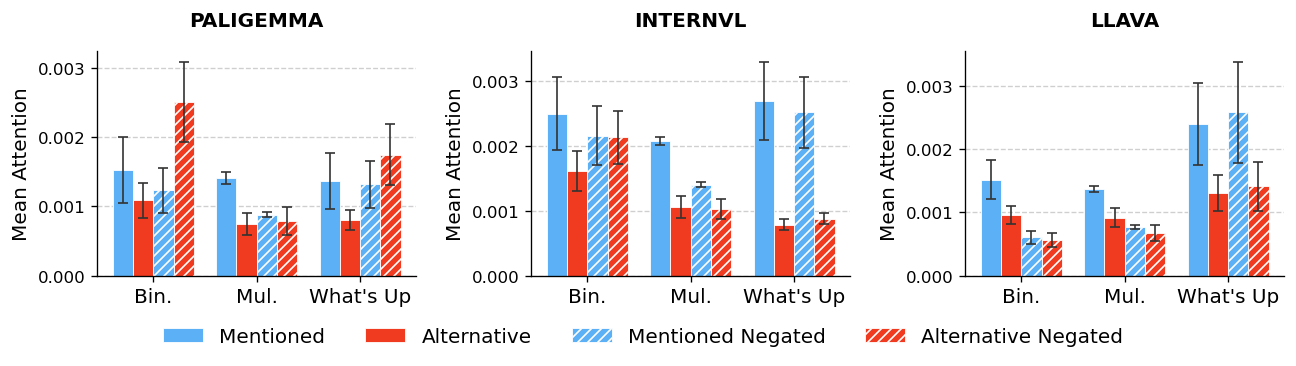}
        \caption{Eight token output.}
        \label{subfig:size_right_large}
    \end{subfigure}
    
    \caption{}
    \label{fig:token_comparison}
\end{figure*}

We repeat our experiments for the three datasets with different values for the maximum number of output tokens generated by each model. For smaller model sizes, we again plot the mean attention to each object in the image (see Figure \ref{fig:bar_graph_results}) over a subset of 20 images from each dataset. Instead of a single token attention we take the mean of the attention from each generated output token. We consider 2, 5 and 8 output token limits and report the attention patterns for each dataset in Figure \ref{fig:token_comparison}

Varying the output tokens does have an effect on attention patterns, for example the attention to the mentioned object in the Binary dataset increases with more output tokens for LlaVa-1.5. PaliGemma2-3B has attention more evenly distributed between mentioned and alternative in the Multary dataset, however, it still is the only model following the expected pattern in each dataset for a range of total generated output tokens. 

\subsection{Analysis of Object Size}
\label{sec:obj_size}
\begin{figure}[htbp]
    \centering
    \begin{subfigure}[b]{0.45\textwidth}
        \centering
        \includegraphics[width=\textwidth]{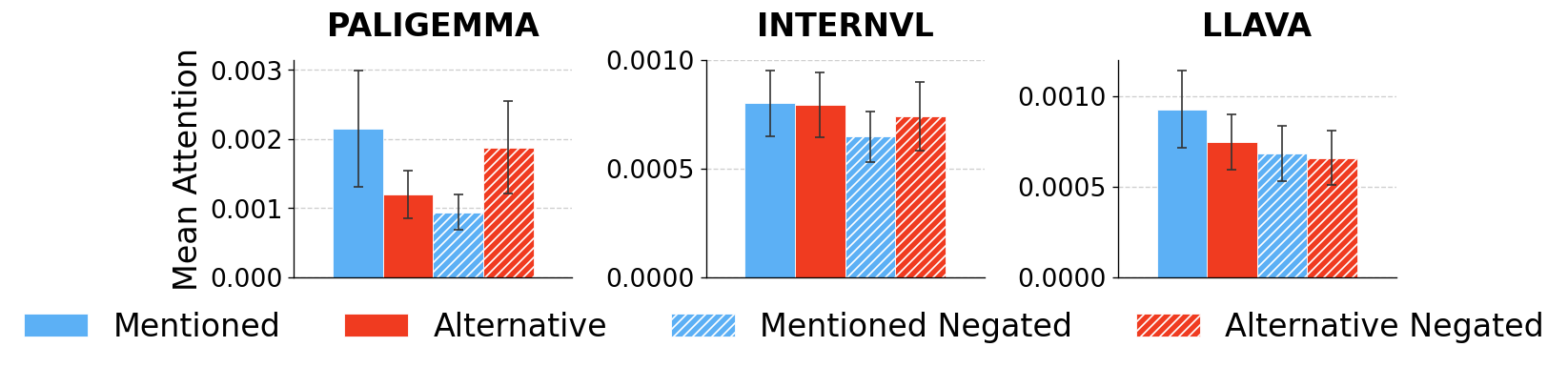}
        \caption{Objects of the same size.}
        \label{subfig:same_size}
    \end{subfigure}
    
    \vspace{0.3cm}
    
    \begin{subfigure}[b]{0.45\textwidth}
        \centering
        \includegraphics[width=\textwidth]{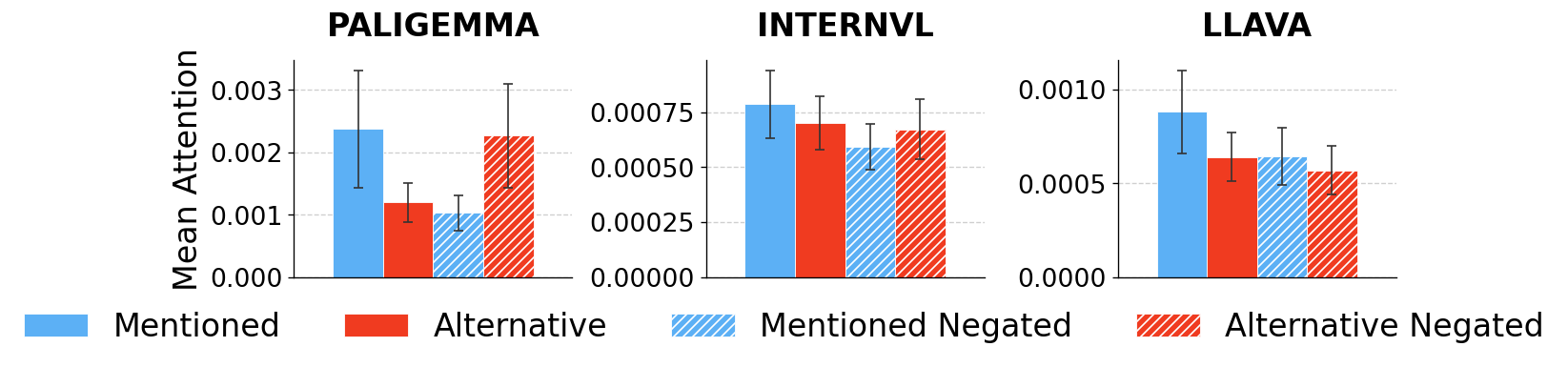}
        \caption{Left object is large, right is small.}
        \label{subfig:size_left_large}
    \end{subfigure}
    
    \vspace{0.3cm}
    
    \begin{subfigure}[b]{0.45\textwidth}
        \centering
        \includegraphics[width=\textwidth]{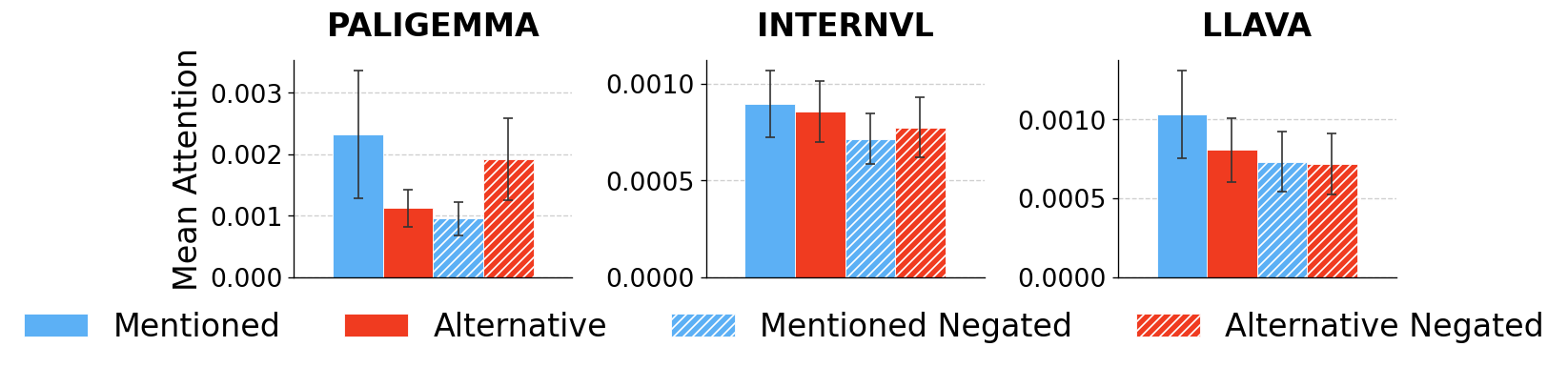}
        \caption{Right object is large, left is small.}
        \label{subfig:size_right_large}
    \end{subfigure}
    
    \caption{Comparison of binary results based on relative object sizes.}
    \label{fig:binary_size_comparison}
\end{figure}

To investigate the effect of object size on attention patterns, we evaluate model responses across three distinct size configurations: images where both objects are identical in size, images where the left object is large and the right object is small, and images where the right object is large and the left object is small. Each set of images contains 20 images, where objects are always the same shape but of differing colours. We repeat our binary experiment as outlined in section \ref{sec:method} for each size configuration over models LLaVA-1.5-7B, Intern-VL-3.5-4B and PaliGemma2-3B. We report mean attention to the mentioned and alternative objects in Figure \ref{fig:binary_size_comparison}.
We find that smaller objects tend to receive more attention per token, likely due to the smaller number of tokens and therefore a smaller denominator for overall attention in the bounding box. However, our overall attention patterns to the mentioned and unmentioned object in affirmative and negated contexts still holds for each of the three size variations tested.  

\subsection{Binary Results}
\label{sec:binary_layer}
\begin{figure*}[t]
    \centering
    \includegraphics[width=.9\textwidth]{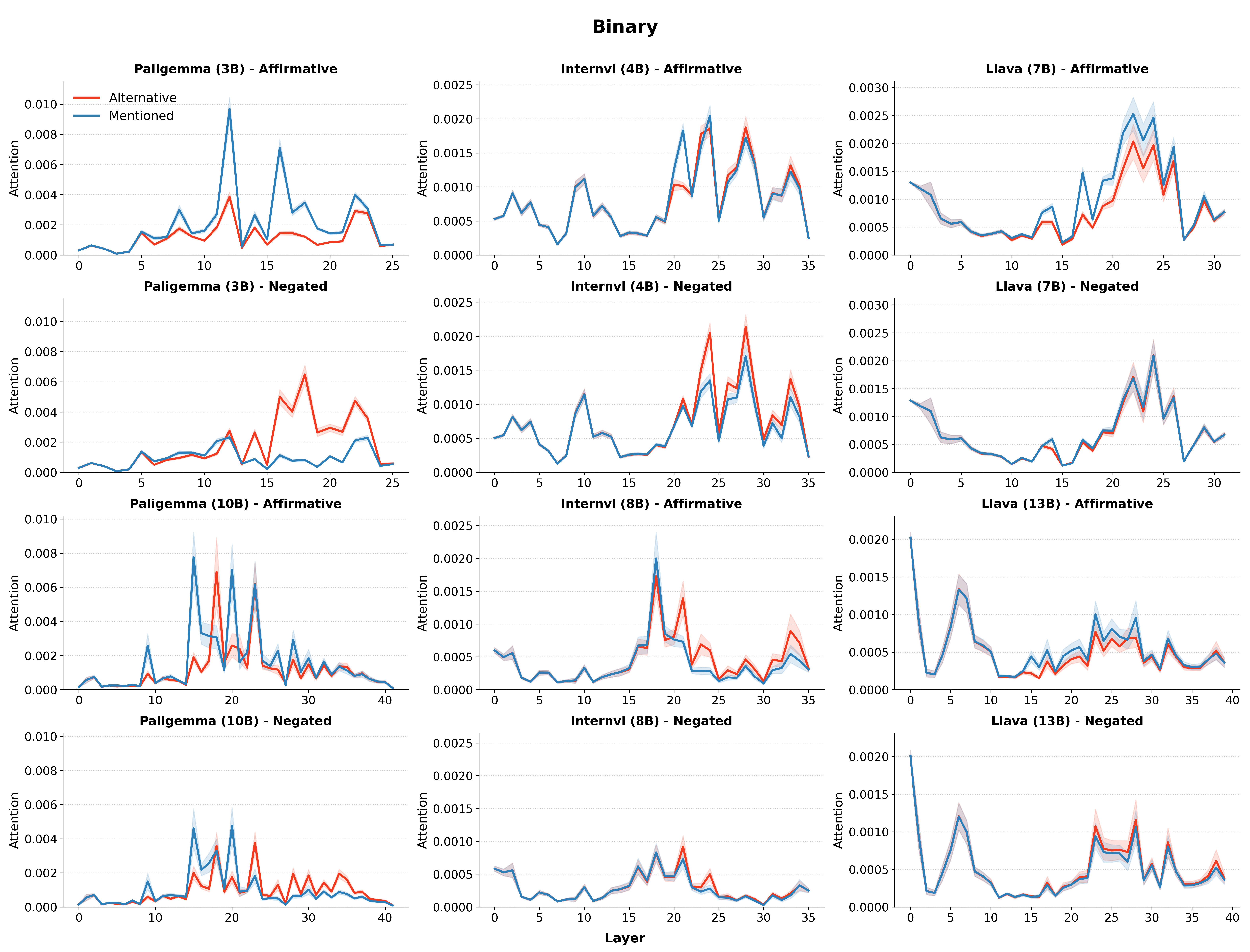}
    \caption{Layer-wise attention for models in the Binary setting.}
    \label{fig:2_shape_results}
\end{figure*}

We visualise the layer-wise attention distribution for all images in the binary dataset. Figure \ref{fig:2_shape_results} shows the layer-wise attention for models in the positive and negated settings. 

\subsection{Multary}
\label{sec:multary_layer}
\begin{figure*}[tbp]
    \centering
    \includegraphics[width=.9\textwidth]{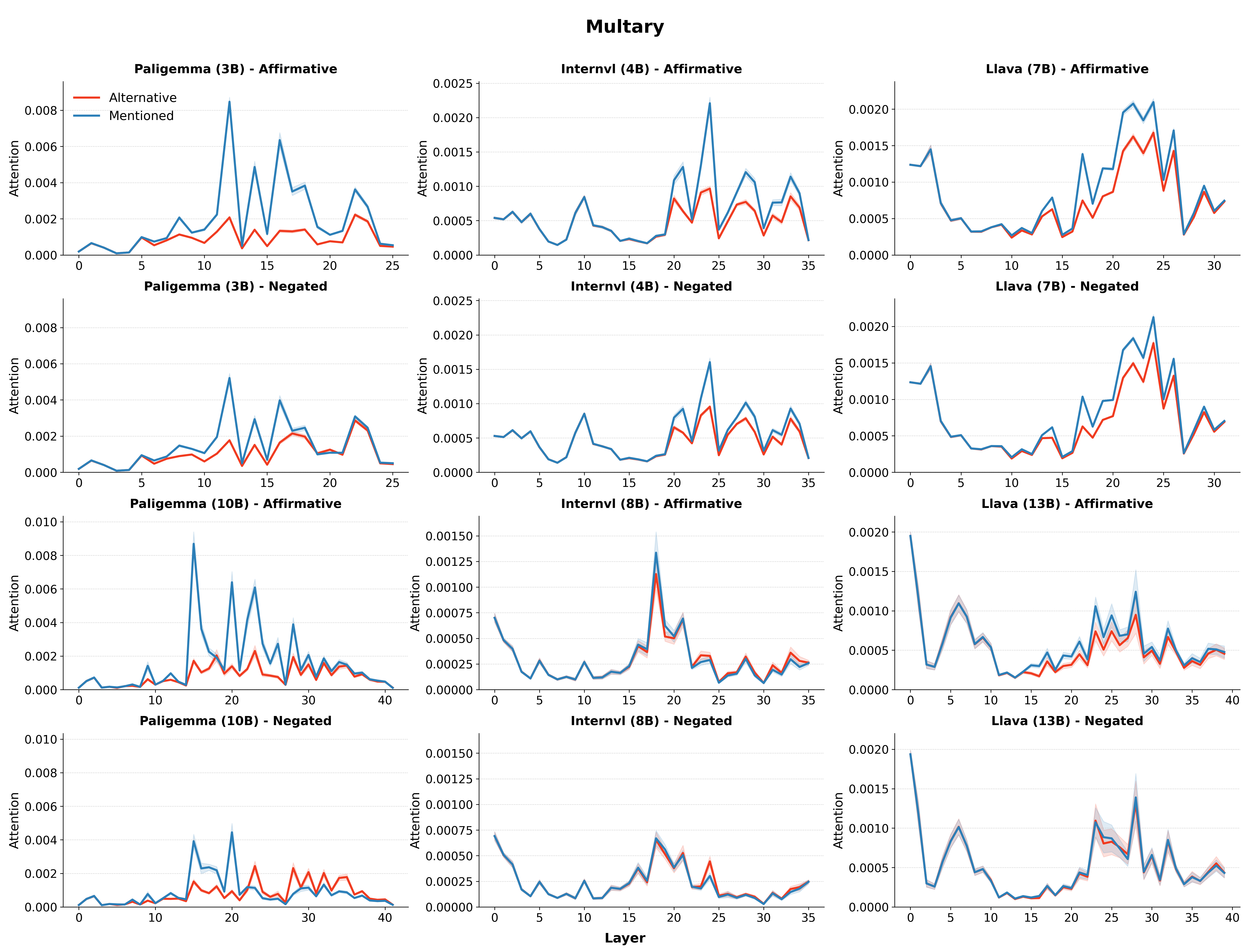}
    \caption{Layer-wise attention for models in the Multary setting.}
    \label{fig:4_shape_results}
\end{figure*}

We report average attention with 95\% confidence intervals at each layer for the three models LLaVA, InternVL and Paligemma2 in Figure \ref{fig:4_shape_results}. Attention to the not mentioned colour is reported as the average attention between the three alternative shapes. 

\subsection{Downstream Tasks Prompt Templates}
\label{sec:prompts}
Questions for the Binary dataset use the following templates:
\begin{quote}
1. Is this statement correct? The object on the \{position\} is \{colour\}. Answer yes or no.

2. Is the object on the \{position\} \{colour\}? Answer yes or no.

3. Does this statement match the image? The object on the \{position\} is \{colour\}. Answer yes or no.

4. Is the object at the \{position\} position \{colour\}? Answer yes or no.

5. Is this true? The object at the \{position\} position is \{colour\}? Answer yes or no.
\end{quote}
The position can either be `left' or `right' which is chosen at random for each image and colour is chosen based on colour of objects within the image. For each prompt template we report mean and standard deviation over 5 seeds for each model.

For the Multary dataset, we use the same prompt templates, however positions can be either `top left', `top right', `bottom left' or `bottom right', for example: `Is this statement correct? The object at the top right is green'. The colour is selected based on the object at the chosen position and false question colours are chosen at random from the three alternative colours.  Questions in the `What'sUp' experiments have the same positions as the Binary dataset, however instead of colour we ask the model for the identity of the object. For example: ``Is the object on the left a chair?'' or ``Is the object on the left not a chair?''.

\subsection{PaliGemma2-3B Layerwise Ablation}
\label{sec:ablation}
\begin{figure}[t]
    \centering
    \includegraphics[width=.45\textwidth]{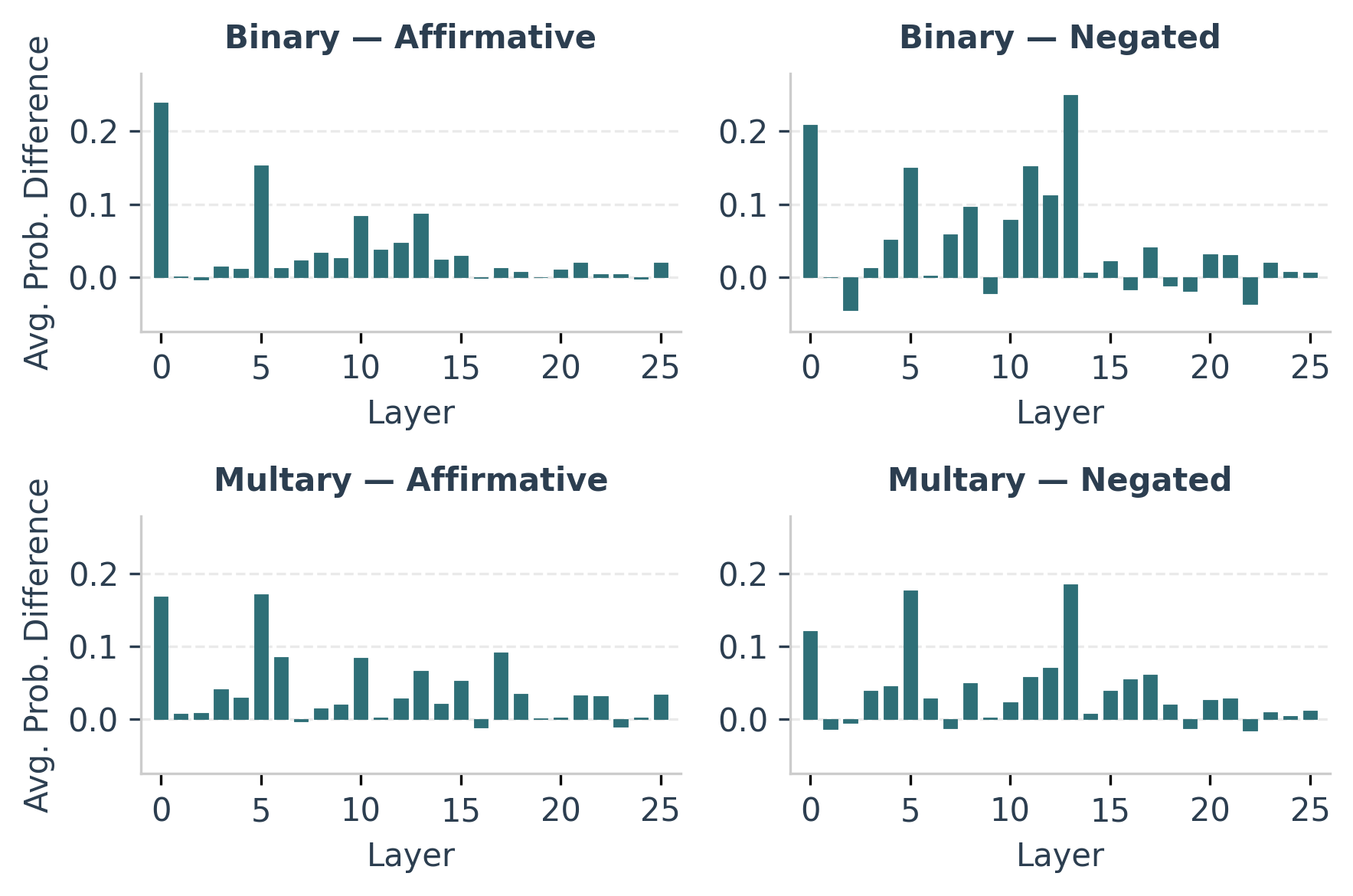}
    \caption{Effect of each layer on correct model output for affirmative and negated prompts across binary and multary datasets. A positive value means the targeted layer caused a decrease in correct prediction.}
    \label{fig:pali_ablation}
\end{figure}

We identify which layers are causally responsible for improving a models predicting performance. We use PaliGemma2-3B as a case study. We filter examples from the binary and multary datasets where the model gets a correct prediction using prompts from the downstream experiment outlined in Section \ref{sec:downstream_full}. We take 40 prompt and image examples for both the binary and the multary datasets. For each layer $l \in [1, L]$, we perform a layer-wise ablation by setting the attention outputs of all heads $h \in [1, H]$ to zero, such that $\mathbf{A}_l^h = \mathbf{0}$ for the targeted layer. We measure the effect of each layer on the model's predictions by calculating the difference in the prediction probability of the correct answer token (`yes' or `no') before and after ablating that layer's attention outputs. For each dataset we use an even amount of prompts for each position (i.e. for binary we use 20 prompts containing the word `left' and 20 prompts with `right'). We report the average probability change over prompts and images in Figure \ref{fig:pali_ablation}. 

We find that earlier layers are more important for correct predictions in both affirmative and negated prompts. The impact of earlier layers is even stronger in negated prompts with layers 13 and earlier having the most effect. While the prompts differ from our object-centric layer-wise plots, we can draw comparisons between the affirmative and negated prompts for each. We hypothesise that since PaliGemma2-3B's attention shifts to the negated object in layers 13+ and layers 13 and earlier have the highest impact on prediction probability, that attention patterns may be informed by negation processing rather than using attention (by simulating the alternative) to process the negation as we see in humans.

\subsection{PaliGemma2-3B Correct and Incorrect Attention}
\label{sec:correct_incorrect}
\begin{figure}[t]
    \centering
    \includegraphics[width=.45\textwidth]{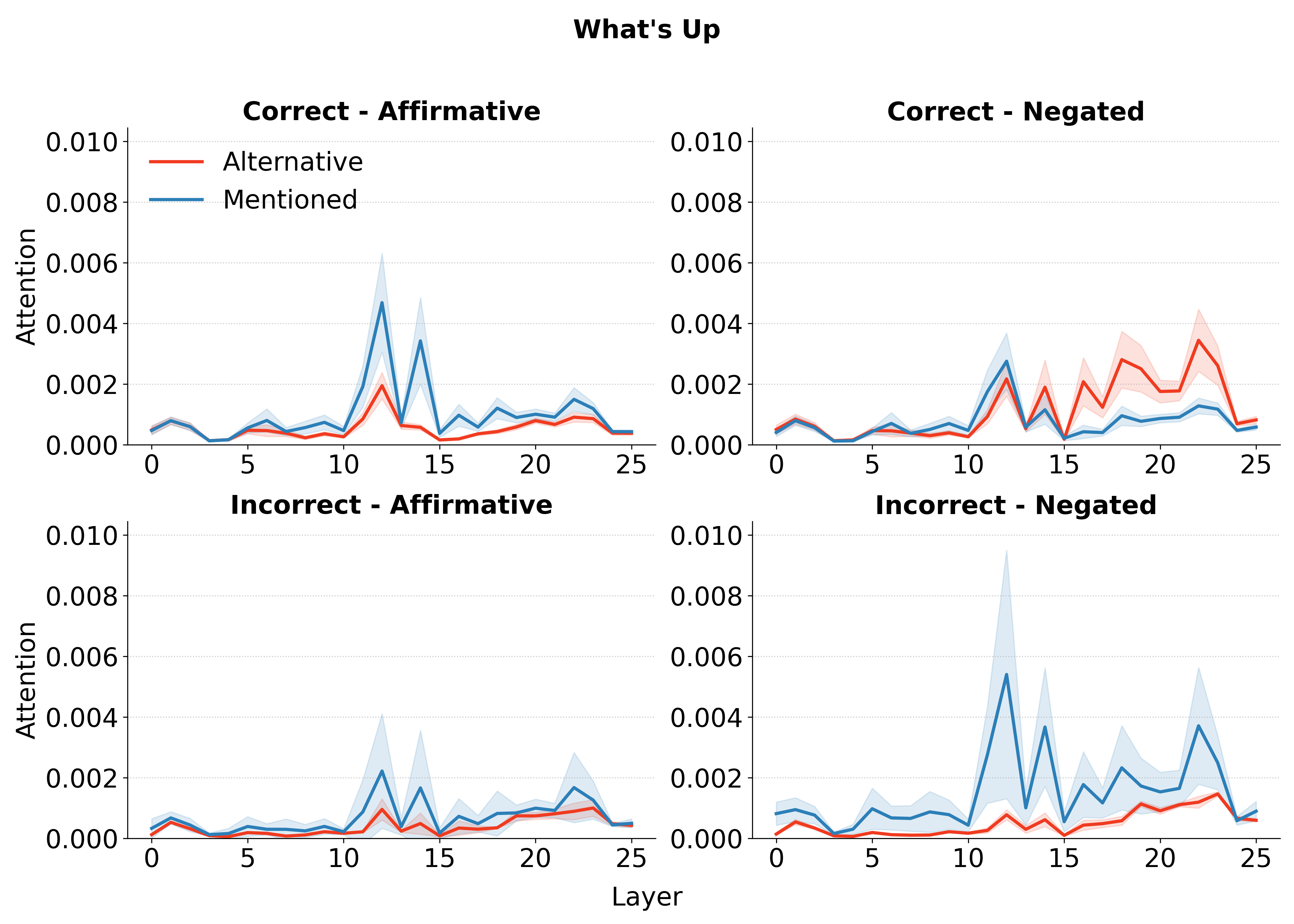}
    \caption{Layer-wise attention plots for PaliGemma2-3B on downstream questions for correct and incorrect responses.}
    \label{fig:correct}
\end{figure}

We plot layer-wise attention on a subset of 20 images from the WhatsUp dataset for PaliGemma2-3b. We use prompt template 1 and compare affirmative true with negated false questions. We find that there is no shift to the negated object when the model has an incorrect prediction, instead we see increased attention to the mentioned object, suggesting that flawed negation processing earlier in the model may be misinforming subsequent attention allocation. Figure \ref{fig:correct} shows the layer-wise breakdown for correct and incorrect predictions.

\subsection{PaliGemma2-3B Head-wise Attention}
\label{sec:headwise}
\begin{figure*}[t]
    \centering
    \includegraphics[width=\textwidth]{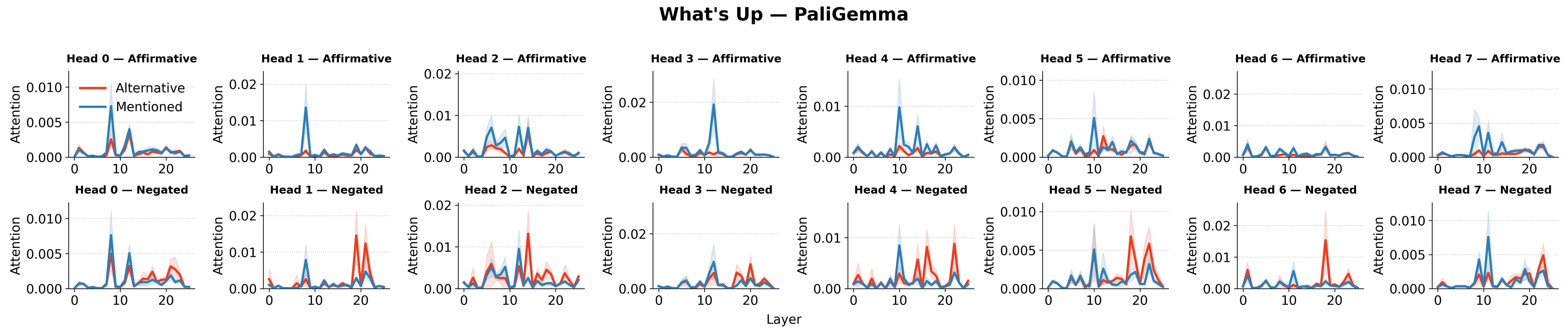}
    \caption{Layerwise attention plots for each head in PaliGemma2-3B on the WhatsUp dataset.}
    \label{fig:layer_heads}
\end{figure*}

We break down the layer-wise attention plots in Figure \ref{fig:whatsup_results} for Paligemma2-3B by attention head. We report the attention to objects in the image from each of the 8 attention heads, shown in Figure \ref{fig:layer_heads}. 
For negated prompts, all heads show some shift in attention to the negated object in later layers however the specific layers with the highest attention varies between heads. This implies the model's understanding isn't handled by a single `negation head' but rather the process is shared between heads. However heads appear to have different roles in this process. Heads 2 and 4 most closely follow the global pattern with attention to the negated object increasing after layer 13. Heads 1 and 6 show spikes in attention to the negated object at isolated later layers, suggesting they act on information processed by earlier layers to focus the models output on the negated object. 

\begin{figure}[t]
    \centering
    \includegraphics[width=.45\textwidth]{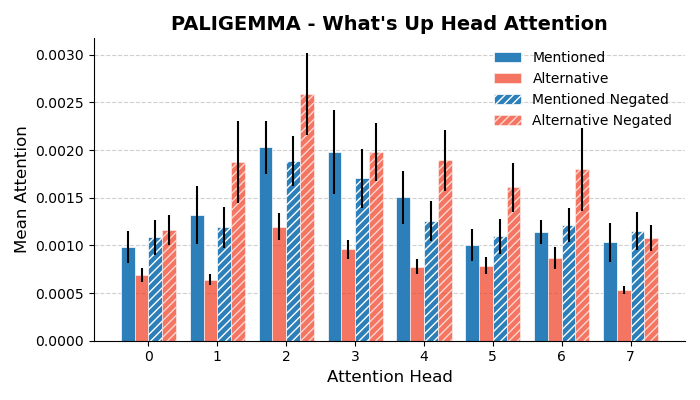}
    \caption{Headwise total attention to objects for PaliGemma2-3B on the WhatsUp dataset}
    \label{fig:total_heads}
\end{figure}

We also plot overall attention to objects in the WhatsUp dataset for each head in PaliGemma2-3B, shown in Figure \ref{fig:total_heads}. We see that all heads except head 7 follow the same global pattern for PaliGemma2 as shown in Figure \ref{fig:bar_graph_results}, again showing there are no specialised `negation heads' and heads mostly share function of shifting attention to the alternative object in the negated case. The specialisation appears mostly layer-based with specific heads attending to the negated objects at different layers. While some heads appear to process negation context at middle layers, others show sharp increases to the negated object only at later layers.

\subsection{Downstream Tasks Results}
\label{sec:downstream_full}
\begin{table*}[t]
\centering
\small 
\resizebox{\textwidth}{!}{%
\begin{tabular}{ll cccc cccc cccc}
\toprule
 & & \multicolumn{4}{c}{\textbf{Binary}} & \multicolumn{4}{c}{\textbf{Multary}} & \multicolumn{4}{c}{\textbf{WhatsUp}} \\
\cmidrule(lr){3-6} \cmidrule(lr){7-10} \cmidrule(lr){11-14}
\textbf{Model} & \textbf{Pr.} & \textbf{A.T} & \textbf{N.T} & \textbf{A.F.} & \textbf{N.F.} & \textbf{A.T} & \textbf{N.T} & \textbf{A.F.} & \textbf{N.F.} & \textbf{A.T} & \textbf{N.T.} & \textbf{A.F.} & \textbf{N.F.} \\ 
\midrule
\multirow{3}{*}{LLaVA-7B}  
 & 1 & 100.0\textsuperscript{0.0} & 19.4\textsuperscript{0.9} & 91.8\textsuperscript{0.6} & 97.8\textsuperscript{0.2} & 95.7\textsuperscript{0.4}  & 1.4\textsuperscript{0.3}  & 41.6\textsuperscript{1.7} & 99.0\textsuperscript{0.3} & 90.7\textsuperscript{2.2} & 2.3\textsuperscript{1.2}  & 33.3\textsuperscript{2.9} & 97.4\textsuperscript{1.2} \\
 & 2 & 100.0\textsuperscript{0.0} & 62.6\textsuperscript{0.8} & 41.7\textsuperscript{1.6} & 100.0\textsuperscript{0.0} & 100.0\textsuperscript{0.0} & 74.3\textsuperscript{2.5} & 12.6\textsuperscript{1.5} & 59.0\textsuperscript{0.9}  & 84.5\textsuperscript{2.3} & 4.9\textsuperscript{1.5}  & 30.8\textsuperscript{1.3} & 93.8\textsuperscript{0.8} \\
 & 3 & 100.0\textsuperscript{0.0} & 0.0\textsuperscript{0.0}  & 82.5\textsuperscript{0.8} & 100.0\textsuperscript{0.0} & 71.9\textsuperscript{1.8}  & 80.3\textsuperscript{1.2} & 78.1\textsuperscript{1.1} & 54.0\textsuperscript{1.5}  & 87.9\textsuperscript{2.3} & 0.3\textsuperscript{0.2}  & 38.6\textsuperscript{3.3} & 99.2\textsuperscript{0.6} \\ 
  & 4 & 100.0\textsuperscript{0.0} & 31.6\textsuperscript{1.0}  & 3.9\textsuperscript{0.5} & 98.8\textsuperscript{0.1} & 100.0\textsuperscript{0.0}  & 71.2\textsuperscript{1.5} & 8.5\textsuperscript{1.5} & 57.6\textsuperscript{1.6}  & 91.8\textsuperscript{2.2} & 8.8\textsuperscript{1.1}  & 23.9\textsuperscript{1.5} & 88.8\textsuperscript{1.6} \\ 
   & 5 & 100.0\textsuperscript{0.0} & 0.0\textsuperscript{0.0}  & 48.0\textsuperscript{1.9} & 100.0\textsuperscript{0.0} & 99.9\textsuperscript{0.1}  & 0.02\textsuperscript{0.0} & 17.9\textsuperscript{1.6} & 100.0\textsuperscript{0.0}  & 88.5\textsuperscript{2.2} & 1.3\textsuperscript{0.6}  & 18.5\textsuperscript{2.2} & 99.0\textsuperscript{0.3} \\ 
  \cmidrule(lr){2-14}
   & \textbf{Avg} & \textbf{100.0}\textsuperscript{0.0} & \textbf{22.7}\textsuperscript{23.77} & \textbf{53.6}\textsuperscript{32.09} & \textbf{99.3}\textsuperscript{0.9} & \textbf{93.5}\textsuperscript{11.18} & \textbf{45.5}\textsuperscript{37.42} & \textbf{31.7}\textsuperscript{26.44} & \textbf{73.9}\textsuperscript{21.41} & \textbf{88.7}\textsuperscript{3.29} & \textbf{3.5}\textsuperscript{3.27}& \textbf{29.0}\textsuperscript{7.53} & \textbf{95.6}\textsuperscript{4.11} \\
   \midrule
  \multirow{3}{*}{LLaVA-13B}  
 & 1 & 100.0\textsuperscript{0.0} & 40.6\textsuperscript{5.4} & 99.6\textsuperscript{0.5} & 100.0\textsuperscript{0.0} & 100.0\textsuperscript{0.0}  & 95.8\textsuperscript{1.0}  & 24.2\textsuperscript{4.0} & 47.2\textsuperscript{3.7} & 86.8\textsuperscript{1.5} & 8.2\textsuperscript{1.5}  & 40.8\textsuperscript{2.9} & 92.6\textsuperscript{0.8} \\
 & 2 & 100.0\textsuperscript{0.0} & 78.6\textsuperscript{1.6} & 77.4\textsuperscript{1.5} & 73.2\textsuperscript{2.6} & 100.0\textsuperscript{0.0} & 59.0\textsuperscript{4.0} & 20.2\textsuperscript{2.3} & 34.0\textsuperscript{2.8}  & 86.6\textsuperscript{1.0} & 72.8\textsuperscript{3.3}  & 55.0\textsuperscript{3.0} & 65.0\textsuperscript{2.1} \\
 & 3 & 100.0\textsuperscript{0.0} & 0.0\textsuperscript{0.0}  & 89.2\textsuperscript{2.1} & 100.0\textsuperscript{0.0} & 100.0\textsuperscript{0.0}  & 94.0\textsuperscript{0.6} & 15.8\textsuperscript{2.2} & 23.8\textsuperscript{4.9}  & 89.2\textsuperscript{1.2} & 0.8\textsuperscript{0.4}  & 43.6\textsuperscript{3.6} & 99.6\textsuperscript{0.5} \\ 
  & 4 & 100.0\textsuperscript{0.0} & 77.6\textsuperscript{2.2}  & 34.2\textsuperscript{3.5} & 69.2\textsuperscript{5.9} & 100.0\textsuperscript{0.0}  & 82.0\textsuperscript{1.7} & 18.0\textsuperscript{2.0} & 54.2\textsuperscript{1.8}  & 92.0\textsuperscript{1.9} & 30.2\textsuperscript{2.5}  & 16.2\textsuperscript{3.7} & 77.8\textsuperscript{1.5} \\ 
   & 5 & 100.0\textsuperscript{0.0} & 0.6\textsuperscript{0.8}  & 95.2\textsuperscript{1.0} & 100.0\textsuperscript{0.0} & 100.0\textsuperscript{0.0}  & 1.8\textsuperscript{1.0} & 44.6\textsuperscript{5.5} & 99.4\textsuperscript{0.5}  & 70.8\textsuperscript{3.0} & 1.6\textsuperscript{1.0}  & 47.8\textsuperscript{3.2} & 96.6\textsuperscript{1.0} \\ 
  \cmidrule(lr){2-14}
 & \textbf{Avg} & \textbf{100.0}\textsuperscript{0.0} & \textbf{39.48}\textsuperscript{34.91} & \textbf{79.12}\textsuperscript{23.75} & \textbf{88.48}\textsuperscript{14.46} & \textbf{100.0}\textsuperscript{0.0} & \textbf{66.52}\textsuperscript{34.98} & \textbf{24.56}\textsuperscript{10.96} & \textbf{51.72}\textsuperscript{26.24} & \textbf{85.08}\textsuperscript{7.64} & \textbf{22.72}\textsuperscript{27.28}& \textbf{40.68}\textsuperscript{13.55} & \textbf{86.32}\textsuperscript{13.09} \\
\midrule
\multirow{3}{*}{InternVL-4B} 
 & 1 & 100.0\textsuperscript{0.0} & 100.0\textsuperscript{0.0} & 100.0\textsuperscript{0.0} & 100.0\textsuperscript{0.0} & 100.0\textsuperscript{0.0} & 95.8\textsuperscript{0.7} & 100.0\textsuperscript{0.0} & 100.0\textsuperscript{0.0} & 68.8\textsuperscript{2.2} & 75.6\textsuperscript{1.3} & 96.8\textsuperscript{0.7} & 82.3\textsuperscript{0.9} \\
 & 2 & 100.0\textsuperscript{0.0} & 100.0\textsuperscript{0.0} & 100.0\textsuperscript{0.0} & 99.1\textsuperscript{0.3}  & 91.6\textsuperscript{0.9}  & 75.0\textsuperscript{0.3} & 100.0\textsuperscript{0.0} & 97.3\textsuperscript{0.3}  & 94.1\textsuperscript{0.9} & 76.2\textsuperscript{1.5} & 68.4\textsuperscript{1.2} & 75.6\textsuperscript{1.4} \\
 & 3 & 100.0\textsuperscript{0.0} & 100.0\textsuperscript{0.0} & 100.0\textsuperscript{0.0} & 100.0\textsuperscript{0.0} & 75.9\textsuperscript{1.1}  & 72.3\textsuperscript{0.8} & 99.8\textsuperscript{0.1}  & 99.3\textsuperscript{0.1}  & 71.7\textsuperscript{2.2} & 90.3\textsuperscript{1.2} & 95.5\textsuperscript{0.9} & 68.8\textsuperscript{1.3} \\ 
  & 4 & 100.0\textsuperscript{0.0} & 100.0\textsuperscript{0.0}  & 100.0\textsuperscript{0.0} & 100.0\textsuperscript{0.0} & 96.1\textsuperscript{0.8}  & 46.3\textsuperscript{1.7} & 100.0\textsuperscript{0.0} & 92.08\textsuperscript{0.4}  & 92.9\textsuperscript{1.0} & 70.7\textsuperscript{1.6}  & 52.0\textsuperscript{2.6} & 74.6\textsuperscript{2.2} \\ 
   & 5 & 100.0\textsuperscript{0.0} & 100.0\textsuperscript{0.0}  & 100.0\textsuperscript{0.0} & 100.0\textsuperscript{0.0} & 99.9\textsuperscript{0.1}  & 96.2\textsuperscript{0.7} & 100.0\textsuperscript{0.0} & 100.0\textsuperscript{0.0}  & 88.0\textsuperscript{1.6} & 0.7\textsuperscript{0.5}  & 39.7\textsuperscript{1.7} & 99.3\textsuperscript{0.94} \\ 
   \cmidrule(lr){2-14}
 & \textbf{Avg} & \textbf{100.0}\textsuperscript{0.0} & \textbf{100.0}\textsuperscript{0.0} & \textbf{100.0}\textsuperscript{0.0} & \textbf{99.8}\textsuperscript{0.40} & \textbf{92.71}\textsuperscript{9.16} & \textbf{77.13}\textsuperscript{18.78} & \textbf{99.95}\textsuperscript{0.11} & \textbf{97.73}\textsuperscript{3.06} & \textbf{83.28}\textsuperscript{11.11} & \textbf{69.86}\textsuperscript{18.33} & \textbf{75.81}\textsuperscript{17.97} & \textbf{78.49}\textsuperscript{7.90} \\
\midrule
\multirow{3}{*}{InternVL-8B} 
 & 1 & 100.0\textsuperscript{0.0} & 100.0\textsuperscript{0.0} & 100.0\textsuperscript{0.0} & 100.0\textsuperscript{0.0} & 97.6\textsuperscript{1.36} & 69.8\textsuperscript{0.75} & 98.2\textsuperscript{1.2} & 95.6\textsuperscript{1.36} & 84.4\textsuperscript{3.0} & 38.6\textsuperscript{2.58} & 76.4\textsuperscript{2.87} & 90.2\textsuperscript{1.47} \\
 & 2 & 100.0\textsuperscript{0.0} & 52.0\textsuperscript{4.3} & 100.0\textsuperscript{0.0} & 94.8\textsuperscript{1.47}  & 100.0\textsuperscript{0.0}  & 90.6\textsuperscript{2.2} & 94.8\textsuperscript{1.9} & 96.8\textsuperscript{1.2}  & 79.2\textsuperscript{3.7} & 3.8\textsuperscript{1.0} & 55.8\textsuperscript{3.3} & 96.6\textsuperscript{1.2} \\
 & 3 & 100.0\textsuperscript{0.0} & 100.0\textsuperscript{0.0} & 100.0\textsuperscript{0.0} & 100.0\textsuperscript{0.0} & 98.6\textsuperscript{0.8}  & 96.2\textsuperscript{0.75} & 76.8\textsuperscript{3.0}  & 14.4\textsuperscript{1.4}  & 62.6\textsuperscript{3.4} & 56.2\textsuperscript{5.7} & 82.6\textsuperscript{2.7} & 83.6\textsuperscript{2.9} \\ 
  & 4 & 100.0\textsuperscript{0.0} & 46.4\textsuperscript{6.83}  & 67.6\textsuperscript{4.2} & 100.0\textsuperscript{0.0} & 100.0\textsuperscript{0.0}  & 63.0\textsuperscript{3.2} & 88.0\textsuperscript{1.3} & 100.0\textsuperscript{0.0}  & 77.0\textsuperscript{3.9} & 13.8\textsuperscript{1.9}  & 35.4\textsuperscript{5.9} & 89.6\textsuperscript{1.4} \\ 
   & 5 & 100.0\textsuperscript{0.0} & 92.8\textsuperscript{1.72}  & 100.0\textsuperscript{0.0} & 100.0\textsuperscript{0.0} & 100.0\textsuperscript{0.0}  & 89.2\textsuperscript{2.0} & 97.8\textsuperscript{1.3} & 100.0\textsuperscript{0.0}  & 89.4\textsuperscript{2.2} & 32.6\textsuperscript{3.1}  & 46.6\textsuperscript{3.6} & 96.2\textsuperscript{1.0} \\ 
   \cmidrule(lr){2-14}
 & \textbf{Avg} & \textbf{100.0}\textsuperscript{0.0} & \textbf{79.20}\textsuperscript{22.73} & \textbf{93.71}\textsuperscript{12.59} & \textbf{98.30}\textsuperscript{3.42} & \textbf{99.24}\textsuperscript{1.21} & \textbf{81.76}\textsuperscript{13.09} & \textbf{91.12}\textsuperscript{8.25} & \textbf{81.36}\textsuperscript{33.54} & \textbf{80.54}\textsuperscript{8.67} & \textbf{28.64}\textsuperscript{19.54} & \textbf{63.92}\textsuperscript{16.28} & \textbf{91.42}\textsuperscript{4.90} \\
\midrule
\multirow{3}{*}{PGemma-3B} 
 & 1 & 100.0\textsuperscript{0.0} & 80.9\textsuperscript{0.7}  & 100.0\textsuperscript{0.0} & 99.6\textsuperscript{0.1} & 100.0\textsuperscript{0.0} & 71.4\textsuperscript{0.7} & 45.2\textsuperscript{1.2}  & 85.2\textsuperscript{0.6}  & 94.1\textsuperscript{0.6} & 22.7\textsuperscript{1.8} & 54.0\textsuperscript{1.2} & 97.4\textsuperscript{0.8} \\
 & 2 & 100.0\textsuperscript{0.0} & 99.6\textsuperscript{0.1} & 100.0\textsuperscript{0.0} & 95.8\textsuperscript{0.1}  & 100.0\textsuperscript{0.0} & 65.1\textsuperscript{1.2} & 84.1\textsuperscript{1.0}  & 99.1\textsuperscript{0.1}  & 81.4\textsuperscript{2.3} & 29.5\textsuperscript{2.0} & 79.0\textsuperscript{1.7} & 98.6\textsuperscript{0.4} \\
 & 3 & 100.0\textsuperscript{0.0} & 26.5\textsuperscript{0.8}  & 100.0\textsuperscript{0.0} & 100.0\textsuperscript{0.0} & 100.0\textsuperscript{0.0} & 89.7\textsuperscript{0.8} & 63.1\textsuperscript{1.2}  & 97.6\textsuperscript{0.2}  & 92.0\textsuperscript{1.6} & 9.5\textsuperscript{0.4} & 30.6\textsuperscript{3.1} & 99.5\textsuperscript{0.3} \\ 

  & 4 & 100.0\textsuperscript{0.0} & 96.5\textsuperscript{0.2}  & 100.0\textsuperscript{0.0} & 92.2\textsuperscript{0.4} & 100.0\textsuperscript{0.0}  & 64.7\textsuperscript{0.7} & 82.9\textsuperscript{0.5} & 98.2\textsuperscript{0.5}  & 90.2\textsuperscript{0.4} & 14.3\textsuperscript{0.8}  & 84.8\textsuperscript{1.1} & 98.6\textsuperscript{0.6} \\ 
   & 5 & 100.0\textsuperscript{0.0} & 77.5\textsuperscript{0.3}  & 100.0\textsuperscript{0.0} & 95.6\textsuperscript{0.3} & 100.0\textsuperscript{0.0}  & 37.2\textsuperscript{1.5} & 49.8\textsuperscript{1.3} & 95.2\textsuperscript{0.7}  & 93.8\textsuperscript{1.2} & 27.8\textsuperscript{2.6}  & 57.2\textsuperscript{0.9} & 94.0\textsuperscript{1.2} \\ 
      \cmidrule(lr){2-14}
 & \textbf{Avg} & \textbf{100.0}\textsuperscript{0.0} & \textbf{76.2}\textsuperscript{26.80} & \textbf{100.0}\textsuperscript{0.0} & \textbf{96.6}\textsuperscript{2.93} &  \textbf{100.0}\textsuperscript{0.0} & \textbf{65.6}\textsuperscript{17.26} & \textbf{65.0}\textsuperscript{16.55} & \textbf{95.1}\textsuperscript{5.22} & \textbf{90.3}\textsuperscript{4.94} & \textbf{20.8}\textsuperscript{8.04} & \textbf{61.1}\textsuperscript{19.87} & \textbf{97.6}\textsuperscript{2.10} \\
 \midrule
\multirow{3}{*}{PGemma-10B} 
 & 1 & 100.0\textsuperscript{0.0} & 39.8\textsuperscript{4.1}  & 100.0\textsuperscript{0.0} & 35.2\textsuperscript{3.8} & 100.0\textsuperscript{0.0} & 87.4\textsuperscript{0.8} & 95.6\textsuperscript{1.7}  & 100.0\textsuperscript{0.0}  & 94.1\textsuperscript{0.6} & 88.6\textsuperscript{2.6} & 43.4\textsuperscript{3.4} & 9.2\textsuperscript{2.3} \\
 & 2 & 100.0\textsuperscript{0.0} & 55.8\textsuperscript{6.7} & 100.0\textsuperscript{0.0} & 94.2\textsuperscript{0.4}  & 100.0\textsuperscript{0.0} & 72.8\textsuperscript{2.9} & 100.0\textsuperscript{0.0}  & 100.0\textsuperscript{0.0}  & 86.2\textsuperscript{1.7} & 69.2\textsuperscript{2.6} & 79.6\textsuperscript{2.4} & 52.2\textsuperscript{2.1} \\
 & 3 & 100.0\textsuperscript{0.0} & 5.2\textsuperscript{1.7}  & 100.0\textsuperscript{0.0} & 48.4\textsuperscript{2.4} & 100.0\textsuperscript{0.0} & 52.2\textsuperscript{1.6} & 99.4\textsuperscript{0.5}  & 100.0\textsuperscript{0.0}  & 82.4\textsuperscript{1.9} & 45.8\textsuperscript{3.2} & 54.0\textsuperscript{3.4} & 54.6\textsuperscript{2.6} \\ 

  & 4 & 100.0\textsuperscript{0.0} & 87.6\textsuperscript{2.9}  & 100.0\textsuperscript{0.0} & 50.8\textsuperscript{6.9} & 100.0\textsuperscript{0.0}  & 74.4\textsuperscript{3.7} & 96.6\textsuperscript{1.5} & 98.6\textsuperscript{1.0}  & 83.4\textsuperscript{2.6} & 74.0\textsuperscript{1.4}  & 70.8\textsuperscript{3.8} & 31.0\textsuperscript{0.9} \\ 
   & 5 & 100.0\textsuperscript{0.0} & 83.6\textsuperscript{3.4}  & 100.0\textsuperscript{0.0} & 0.0\textsuperscript{0.0} & 100.0\textsuperscript{0.0}  & 69.0\textsuperscript{4.8} & 90.8\textsuperscript{3.5} & 6.6\textsuperscript{1.9}  & 87.6\textsuperscript{2.0} & 99.2\textsuperscript{0.4}  & 47.4\textsuperscript{4.5} & 1.2\textsuperscript{0.4} \\ 
      \cmidrule(lr){2-14}
 & \textbf{Avg} & \textbf{100.0}\textsuperscript{0.0} & \textbf{54.4}\textsuperscript{30.58} & \textbf{100.0}\textsuperscript{0.0} & \textbf{45.72}\textsuperscript{30.5} &  \textbf{100.0}\textsuperscript{0.0} & \textbf{71.16}\textsuperscript{11.75} & \textbf{96.48}\textsuperscript{3.79} & \textbf{81.04}\textsuperscript{37.24} & \textbf{86.2}\textsuperscript{3.84} & \textbf{75.36}\textsuperscript{18.35} & \textbf{59.04}\textsuperscript{14.35} & \textbf{29.64}\textsuperscript{21.81} \\
\bottomrule
\end{tabular}%
}\caption{Downstream Accuracy (Mean \% $\pm$ SD) across datasets for each prompt template}
\label{tab:downstream_results_full}
\end{table*}
The high variance between different prompt styles for models highlights the prompt sensitivity of these models. LLaVA models in particular are very sensitive to prompts especially in negated-true and affirmative false settings with ranges of 0\% to 62.6\% accuracy and 3.9\% to 91.8\% for LLaVA-1.5-7B and 0\% to 78.6\% and 34.2\% to 99.6\% for LLaVA-1.5-13B on the Binary dataset . Models tend to have a larger range on negated-true questions  suggesting that the presence of the word `not' may help prime the models to predict `no', as performance is generally more stable for negated-false questions. 

\section{GPU hours used}
Total GPU hours used was approximately 100 for all experiments.

\end{document}